\documentclass[10pt, reqno, letterpaper, twoside]{amsart}

\usepackage{microtype}
\usepackage{graphicx}
\usepackage{booktabs} 
\usepackage{hyperref}
\pdfpageattr {/Group << /S /Transparency /I true /CS /DeviceRGB>>}

\usepackage[usenames,dvipsnames,svgnames,table]{xcolor}
\usepackage{amsthm, amssymb, bbm, bm, mathtools}
\usepackage{enumerate,enumitem}
\usepackage{graphicx}
\usepackage{float}
\usepackage{wrapfig, subcaption}
\usepackage[margin=0cm,font=small]{caption}
\usepackage[titletoc, toc]{appendix}
\usepackage{microtype}
\usepackage{tabularx}
\usepackage{booktabs,colortbl}
\usepackage{nicefrac}
\usepackage{blindtext}
\usepackage{setspace}
\usepackage{marginnote}
\usepackage{multirow}
\usepackage{csquotes}
\usepackage[scr=boondoxo]{mathalfa}

\usepackage[boxruled, vlined, linesnumbered]{algorithm2e}
\SetAlFnt{\small}
\SetAlCapFnt{\small}
\SetAlCapNameFnt{\small}
\usepackage{algorithmic}

\let\oldnl\nl
\newcommand{\nonl}{\renewcommand{\nl}{\let\nl\oldnl}}

\usepackage[hyperpageref]{backref}
\colorlet{mylinkcolor}{Blue}
\hypersetup{
colorlinks=false,
linkcolor=mylinkcolor, citecolor=mylinkcolor,
filecolor=magenta, urlcolor=mylinkcolor,
}
\usepackage{url, bookmark}

\makeatletter
\def\th@plain{%
  \thm@notefont{}
  \itshape 
}
\def\th@definition{%
  \thm@notefont{}
  \normalfont 
}
\makeatother

\usepackage{array}
\makeatletter
\g@addto@macro{\endtabular}{\rowfont{}}
\makeatother
\newcommand{\rowfonttype}{}
\newcommand{\rowfont}[1]{
   \gdef\rowfonttype{#1}#1%
}
\newcolumntype{L}{>{\rowfonttype}l}

\usepackage[capitalize,nameinlink]{cleveref}
\crefname{definition}{Def.}{Defs.}
\crefname{appendix}{Appendix}{Appendices}
\crefformat{equation}{(#2#1#3)}
\crefmultiformat{equation}{(#2#1#3)}%
{ and~(#2#1#3)}{, (#2#1#3)}{ and~(#2#1#3)}

\crefname{section}{Sec.}{Secs.}
\crefname{subsection}{Sec.}{Secs.}
\crefname{subsubsection}{Sec.}{Secs.}

\newcommand{\reals}{\mathbb{R}}

\newcommand{\tbf}[1]{\textbf{#1}}

\DeclarePairedDelimiter\norm{\lVert}{\rVert}

\newcommand{\ag}[1]{\ensuremath \left\langle#1\right\rangle}

\DeclareMathOperator*{\argmin}{arg\;min}
\DeclareMathOperator*{\argmax}{arg\;max}

\newcommand{\aeq}[1]{\begin{align} #1 \end{align}}

\newcommand{\beq}[1]{\begin{equation}#1\end{equation}}

\newcommand{\trm}[1]{\mathrm{#1}}

\newcommand{\bmat}[1]{\begin{bmatrix}#1\end{bmatrix}}

\providecommand\f[2]{\ensuremath \frac{#1}{#2}}
\providecommand\rbrac[1]{\ensuremath \left(#1\right)}

\providecommand\bigsqbrac[1]{\ensuremath \big[#1 \big]}
\providecommand\Bigsqbrac[1]{\ensuremath \Big[#1 \Big]}

\providecommand\cbrac[1]{\ensuremath \left\{#1\right\}}

\newtheorem{theorem}{Theorem}

\newtheorem{lemma}[theorem]{Lemma}

\theoremstyle{plain}
\newtheorem{definition}[theorem]{Definition}

\newtheorem{remark}[theorem]{Remark}

\newcommand{\E}{\mathbb{E}}

\providecommand{\ones}{\mathbbm{1}}

\definecolor{color_skyblue}{rgb}{0.01,0.39,0.75}

\renewcommand{\th}{\theta}

\newcommand{\e}{\epsilon}

\renewcommand{\l}{\lambda}

\providecommand{\OO}{\mathcal{O}}

\def \MM {\mathcal{M}}

\def \XX {\mathcal{X}}

\def \DD {\mathcal{D}}

\def \OO {\mathcal{O}}

\def \PP {\mathcal{P}}

\newcommand{\bR}{\mathbb{R}}

\newcommand{\ignore}[1]{}

\def \kh {\widehat{\k}}
\def \yh {\widehat{y}}

\def \lce {\ell_{\trm{CE}}}

\def \lw {\ell_{\trm{W}}}

\def \T {\pi}

\def \k {\kappa}

\def \rkk {\reals_+^{K \times K}}
\def \rk {\reals^{K}}

\def \cp {C^{\: \odot p}}

\def \kkp {{k,k'}}

\def \wp {\trm{W}_p}
\def \wpp {\trm{W}_p^p}
\def \lwpp {^\l \trm{W}_p^p}

\def \H {\trm{H}}
\def \Ct {CIFAR-10\xspace}
\def \Ch {CIFAR-100\xspace}
\def \Ti {Tiny-ImagetNet\xspace}

\def \ce {\trm{\textbf{CE}}}

\def \rce {\trm{\textbf{PGD}}}
\def \rw {\bf \trm{\textbf{WPGD}}}

\def \allcnnl {\textrm{All-CNN}}

\def \wrnst {\textrm{W-16-10}}
\def \wrnte {\textrm{W-28-10}}
\def \wrnft {\textrm{W-40-10}}

\graphicspath{{./}}

\newcommand{\rsz}[1]{\small{#1}}

\title{Directional Adversarial Training for Cost Sensitive Deep Learning Classification Applications}

\author[Terzi, Susto and Chaudhari]{
Matteo Terzi$^1$, Gian Antonio Susto$^{1,2}$ Pratik Chaudhari$^3$
}

\begin{document}
\maketitle

{
\vspace*{-0.25in}
\small
\noindent $^1$ Human Inspired Technology Center, University of Padova.\\
$^2$ Department of Information Engineering, University of Padova.\\
\noindent $^3$ University of Pennsylvania.\\
Email:\ \href{mailto:matteo.terzi@phd.unipd.it}{matteo.terzi@phd.unipd.it},\href{mailto:gianantonio.susto@dei.unipd.it}{gianantonio.susto@dei.unipd.it},
\href{mailto:pratikac@seas.upenn.edu}{pratikac@seas.upenn.edu}\\
}

\vskip 0.3in



{
\small
\noindent \textbf{Abstract:}
In many real-world applications of Machine Learning it is of paramount importance not only to provide accurate predictions, but also to ensure certain levels of robustness. Adversarial Training is a training procedure aiming at providing models that are robust to worst-case perturbations around predefined points. 
Unfortunately, one of the main issues in adversarial training is that robustness w.r.t. gradient-based attackers is always achieved at the cost of prediction accuracy. In this paper, a new algorithm, called Wasserstein Projected Gradient Descent (WPGD), for adversarial training is proposed. WPGD provides a simple way to obtain cost-sensitive robustness, resulting in a finer control of the robustness-accuracy trade-off. Moreover, WPGD solves an optimal transport problem on the output space of the network and it can efficiently discover directions where robustness is required, allowing to control the directional trade-off between accuracy and robustness. The proposed WPGD is validated in this work on image recognition tasks with different benchmark datasets and architectures. Moreover, real world-like datasets are often unbalanced: this paper shows that when dealing with such type of datasets, the performance of adversarial training are mainly affected in term of standard accuracy.

}

\section{Introduction}
\label{s:intro}
Recent advancements in Deep Learning have lead to several breakthrough applications in many fields, like Computer Vision~\cite{krizhevsky2012imagenet}, Health-care~\cite{esteva2019guide}, Industry 4.0~\cite{maggipinto2018computer, qi2018rotor}, Natural Language Processing~\cite{young2018recent}, Speech Recognition~\cite{nassif2019speech} and Transportation~\cite{gao2018ima}. A crucial requirement for many applications in these fields, is to have  models that do not have unexpected behaviors. However, Deep neural networks (DNNs), under some circumstances do not satisfy this property.

Probably the main alarming behavior of DNNs~\cite{beke2019learning, lecun2015deep} for classification tasks is that they are susceptible to adversarial perturbations, i.e., for example, in the context of Computer Vision, modifications to the input image that although imperceptible to the human eye cause the network to misclassify, confidently, the image~\cite{szegedy2013intriguing}. These perturbations are easy to synthesize and they may even generalize across different networks~\cite{moosavi2017universal}. This suggests surprising vulnerabilities in these state-of-the-art classifiers and it has resulted in a flurry of activities towards understanding this phenomenon~\cite{fawzi2018analysis,schmidt2018adversarially}, building robustness and defenses against it~\cite{goodfellow2014explaining,madry2017towards}, as also discovering new attacks~\cite{athalye2018obfuscated,carlini2017adversarial,papernot2016transferability,papernot2016practical}. Adversarial robustness is fundamental in many real-world applications; in important applications like autonomous driving~\cite{qayyum2019securing} and predictive maintenance~\cite{susto2014machine}, errors and faults have different priorities and importance: for example, in autonomous driving, if a recognition system of an autonomous car misclassifies a cat as a dog there should be reasonably no damage, while, if a human is misclassified, that could lead to dramatic consequences. 

Adversarial robustness is here defined as the accuracy of a given model evaluated in the worst-case input around a prescribed neighbourhood. More informally, it can be considered as the accuracy of the models in worst-case scenarios.
In this context, the most common and effective approach to enable robustness to adversarial examples in DNNs is \emph{Adversarial Training}~\cite{madry2017towards}, whose idea is to train a model with these worst-case examples, called \emph{adversarial examples} instead of using clean data, ie. data measured either without error or with negligible error. 
Thus, it is training procedure belonging to the class of \emph{minimax problems}~\cite{rafique2018nonconvex}, in which a inner loop finds the worst-case data point $x^\star$ trough gradient ascent and the outer loop minimizes the target loss on $x^\star$.

Unfortunately, adversarial robustness comes at the price of lower classification accuracy on clean data: this trade-off has been demonstrated by various analyses~\cite{fawzi2018adversarial, tsipras2018there}. As argued above, an adversarially robust classifier with low accuracy is unlikely to be used in practical applications require both.
Although much efforts has been devoted to theoretically understand robustness, its practical consequences in industrial applications received few attention from the literature~\cite{ibitoye2019analyzing}. 

The present work aims at addressing the aforementioned issues with the following contributions:
\begin{itemize}
    \item it is shown that the quantitative and qualitative difference between robust and standard models correlates with the visual metric of classes, ie. it is aligned with the human notion of distance between classes. Adversarially trained networks learn to (mostly) ignore fine-grained classification and confuse classes with samples that are close to the decision boundaries. This result is corroborated by~\cite{Moosavi-Dezfooli:2018aa} where it is shown that adversarial training leads to boundaries with low curvature;
    \item it is shown that robust models are less confident in their predictions than standard models are;
    \item inspired by the previous observation, Wasserstein Projected Gradient Descent (WPGD), an algorithm for adversarial training of deep networks, is presented here. WPGD improves the efficiency of the inner loop in gradient-based defenses such as Projected Gradient Descent (PGD). WPGD formulates an optimal transport problem on the label space with the underlying metric given by the distances of the classification boundaries between classes. This metric guides the search for adversarial perturbations towards classes that are visually dissimilar. 
    It is shown that training deep networks using WPGD is effective in shaping boundaries to maintain direction robustness where required will maintaining accuracy on similar classes.


\end{itemize}

\noindent Moreover, it is worth noting that, although the experiments in this work regard image recognition tasks, the WPGD framework can be easily extended to other types of data such as time-series.

The rest of this paper is organized as follows. In~\cref{s:building_blocks} the building blocks of the proposed approach, estimating the distance to the boundaries and optimal transport, are presented, while properties of adversarial training  are discussed in~\cref{s:properties_adversarial}. In \cref{s:wpgd} the WPGD algorithm is introduced and experimental results on MNIST~\cite{lecun1998gradient}, CIFAR-10 and Tiny~Imagenet datasets for different deep networks are reported. Related works and discussion are provided in~\cref{s:related_work} and~\cref{s:discussion} respectively. 

\section{Notation and building blocks}
\label{s:building_blocks}

This section describes the notation and the main building blocks of the approach presented in this work.

\tbf{Notation:} Let $\th \in \reals^d$ denote the parameters of a neural network. Input images are denoted by $X = \cbrac{x_i: i \leq N}$ with pixel intensities normalized to lie between $[0, 1]$. Given an image $x$, let $\k(x) \in \cbrac{1, \ldots, K}$ be its ground-truth label, the one-hot encoding of $\k(x)$ is denoted by $y(x)$. The normalized probability distribution over the classes as predicted by the network is denoted by $\yh(x) \in \rk$, here $\yh(x)_k$ denotes its $k^{\trm{th}}$ entry and $\kh(x) = \argmax_k \yh(x)_k$ is the predicted class. The cross-entropy loss can then be written as
\beq{
    \lce(\th;\ x) = -\log \yh(x)_{\k(x)}
    \label{eq:ellce}
}
and training a network involves minimizing the average loss, ie. $\argmin_{\th}\ \E_{x \sim X} \bigsqbrac{\lce(\th;\ x)}$.

The training dataset is represented with $\DD= \{\bf{x},\bf{y}\}$, where $\tbf{x} = \{x_i\}_{i=1}^N$ and $\tbf{y} = \{y_i\}_{i=1}^N$ are, respectively, a set of randomly sampled data point and their corresponding labels generated from a unknown distribution $p_\psi(x,y)$, parametrized by $\psi$.  
In lieu of minimizing the expected loss over the training data, adversarial training solves
\beq{
    \min_\th\ \E_{X}\ \Bigsqbrac{\max_{x'\ \in\ \MM(x)}\ \lce(x';\ \th)};
    \label{eq:adversarial_training}
}
this is a saddle point problem where, at each iteration, candidate images $x'$ are chosen from a set $\MM(x)$ (or a manifold). This has been a successful approach to training neural networks robustly w.r.t. adversarial perturbations, see~\cite{madry2017towards,kolter2017provable,sinha2018certifying,kannan2018adversarial}. In this paper only  $\MM(x) = \MM_\infty(x) = \cbrac{x':\ \norm{x'-x}_\infty \leq \e}$, the infinity-norm ball around $x$, is considered to obtain an algorithm based on PGD~\cite{boyd2004convex},~\cite{madry2017towards}. 

It is remarked that the theoretical properties described in the following are generally applied to general setting and not only Euclidean perturbations.
In this paper it is distinguished between \emph{natural error} (NE) and \emph{adversarial error} (AE) as the errors obtained with natural images and with adversarial images, respectively.
In the following only $\ell_\infty$ is used for perturbations in all the experiments regarding real datasets while $\ell_2$\footnote{The reason for using $\ell_2$ instead of $\ell_\infty$ is simply to ease visualization of the impact of adversarial training.} for perturbations in the synthetic example of~\cref{ss:boundaries}. 


\section{Properties of adversarial training}
\label{s:adv_tra}
\label{s:properties_adversarial}

In this Section some effects and properties of adversarial training on various aspects are reported. Such aspects are:
\begin{itemize}
    \item the qualitative and quantitative description of classification errors, measured by the accuracy gap (\cref{ss:effects});
    \item unbalanced classification problems (\cref{ss:unbalance});
      \item the characterization of output confidence (~\cref{subsec:entropy_reg});
    \item the characterization of boundaries (\cref{ss:boundaries}).
\end{itemize}
The aforementioned effects are supported by experiments reported in this Section.  Moreover, it is shown in~\cref{subsec:entropy_reg}  that an entropic regularization help in obtaining robustness. 

The properties and effects of adversarial training reported here have motivated WPGD that will be presented in the following Section.

\subsection{Accuracy gap}
\label{ss:effects}

In order to ease the understanding of the results on this Section, the notion of \emph{accuracy gap} is defined as the following:
\begin{definition}
Let $C_{pgd}$ and $C_{ce}$ be the confusion matrices of robust and standard models, respectively. The \emph{accuracy gap} $G$ is defined as the absolute difference between the confusion matrices:
$$G = |C_{pgd} - C_{ce}|$$
\end{definition}

Although it is known that robustness is obtained at cost of accuracy~\cite{madry2017towards,tsipras2018there}, it is not still clear in the literature whether this gap can be mitigated\footnote{On MNIST dataset, high capacity networks reduce the accuracy gap to near zero. However, in more complex datasets, such as \Ct, this gap exists even with very large networks.}. In this work a first step into tackling this problem is taken by studying how errors are distributed between images and classes: it is shown in the following that mis-classification errors are distributed following the \emph{visual metric}, meaning that robust networks tend to destroy fine-grained classification.  Qualitatively, the visual metric is a distance between classes that can be easily interpreted by humans.  One approach for defining such visual metric is to employ the distance from boundaries of a deep neural network: in fact,~\cite{saxe2019mathematical} showed that NNs learn representations that are well-aligned with our idea of visual similarity.


Due to high-dimensionality of input, obtaining a good approximation of the visual metric is not easily feasible. However, it can be replaced by the semantic metric provided by WordNet~\cite{miller1995wordnet}, which is a good proxy for the visual metric as also showed by~\cite{Deng_2010}. For MNIST, it used a linear classifier on the input pixels whereby the boundaries can computeed accurately. 

\begin{figure}[H]
\centering
\captionsetup[subfigure]{justification=centering}
\begin{subfigure}[t]{0.48 \columnwidth}
    \centering
    \includegraphics[width=\textwidth]{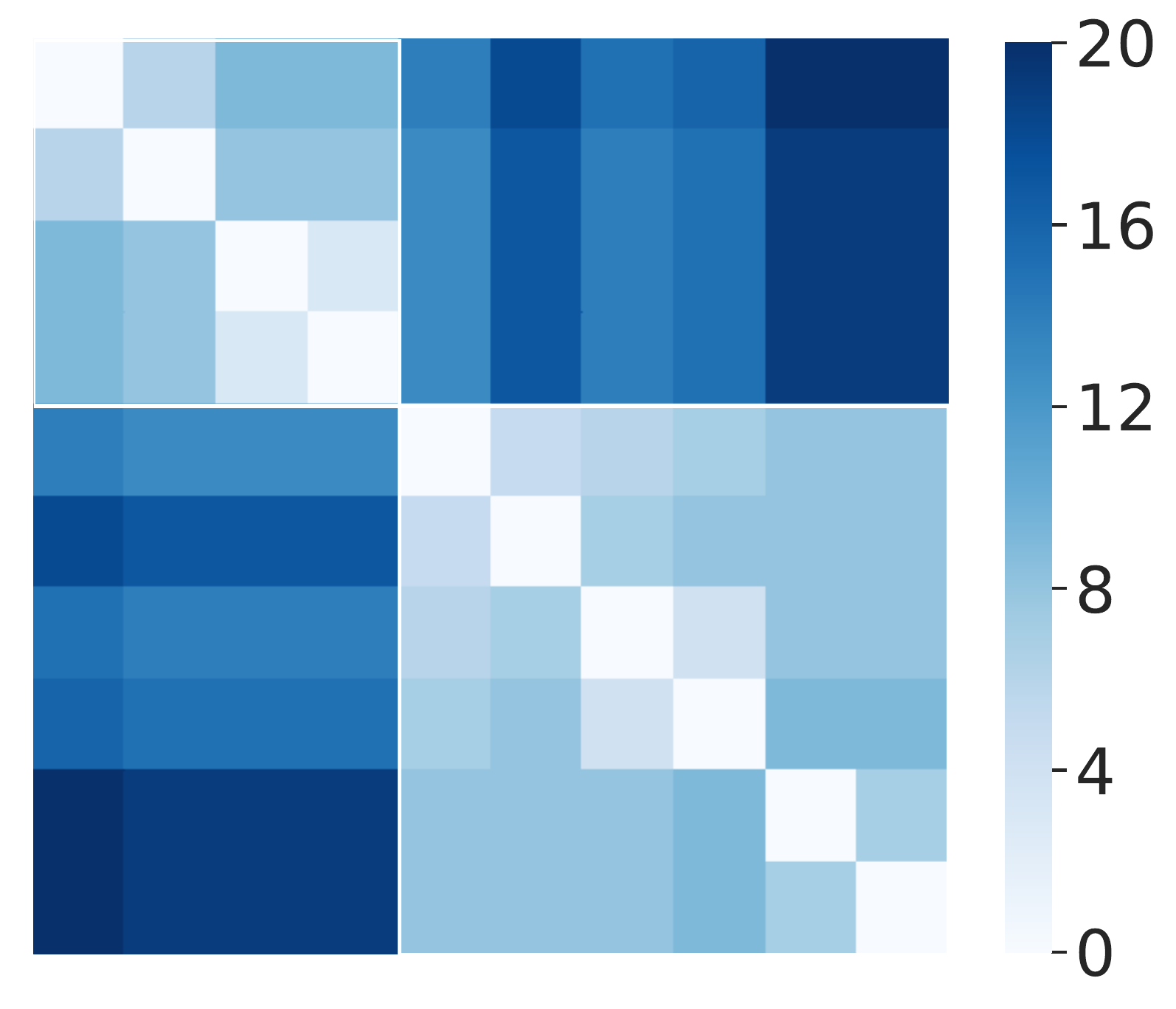}
    \caption{}
    \label{fig:metric_c10}
\end{subfigure}
\begin{subfigure}[t]{0.5 \columnwidth}
    \centering
    \includegraphics[width=\textwidth]{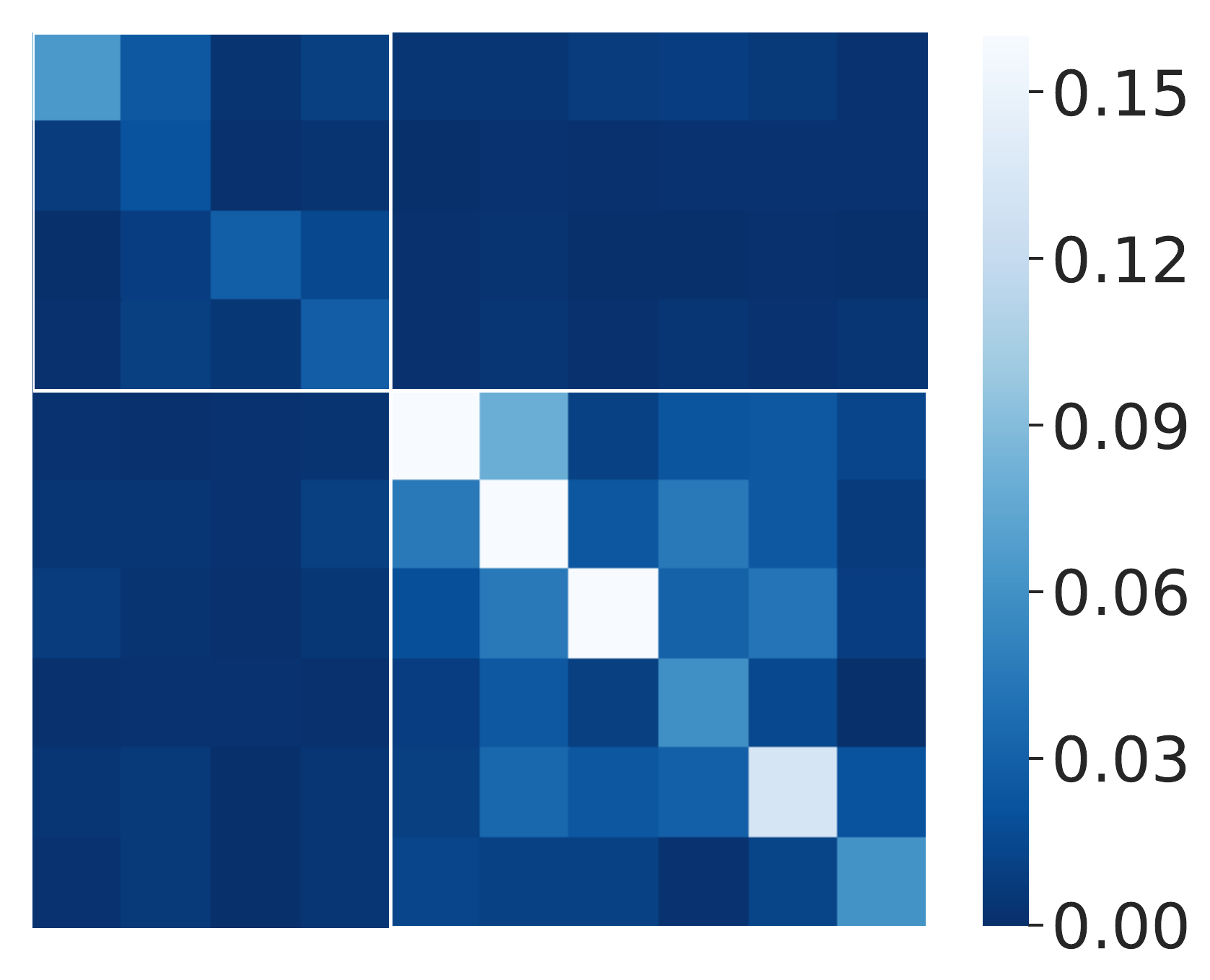}
    \caption{}
    \label{fig:accuracy_gap_wrn_8}
\end{subfigure}
\caption{\Ct dataset. In panel (a) it is reported a matrix of pairwise distances between classes.
Classes are (top to bottom): airplane, car, bird, cat, deer, dog, frog, horse, ship, truck. Panel (b) shows the accuracy gap between a Wide Residual Network \cite{zagoruyko2016wide} trained using PGD and one trained with the standard cross-entropy loss.}
\label{fig:c10_accuracy_gap}
\end{figure}
\begin{figure*}[htpb]
\centering
\captionsetup[subfigure]{justification=centering}
\begin{subfigure}[t]{0.3\textwidth}
        \centering
        \includegraphics[width=\textwidth]{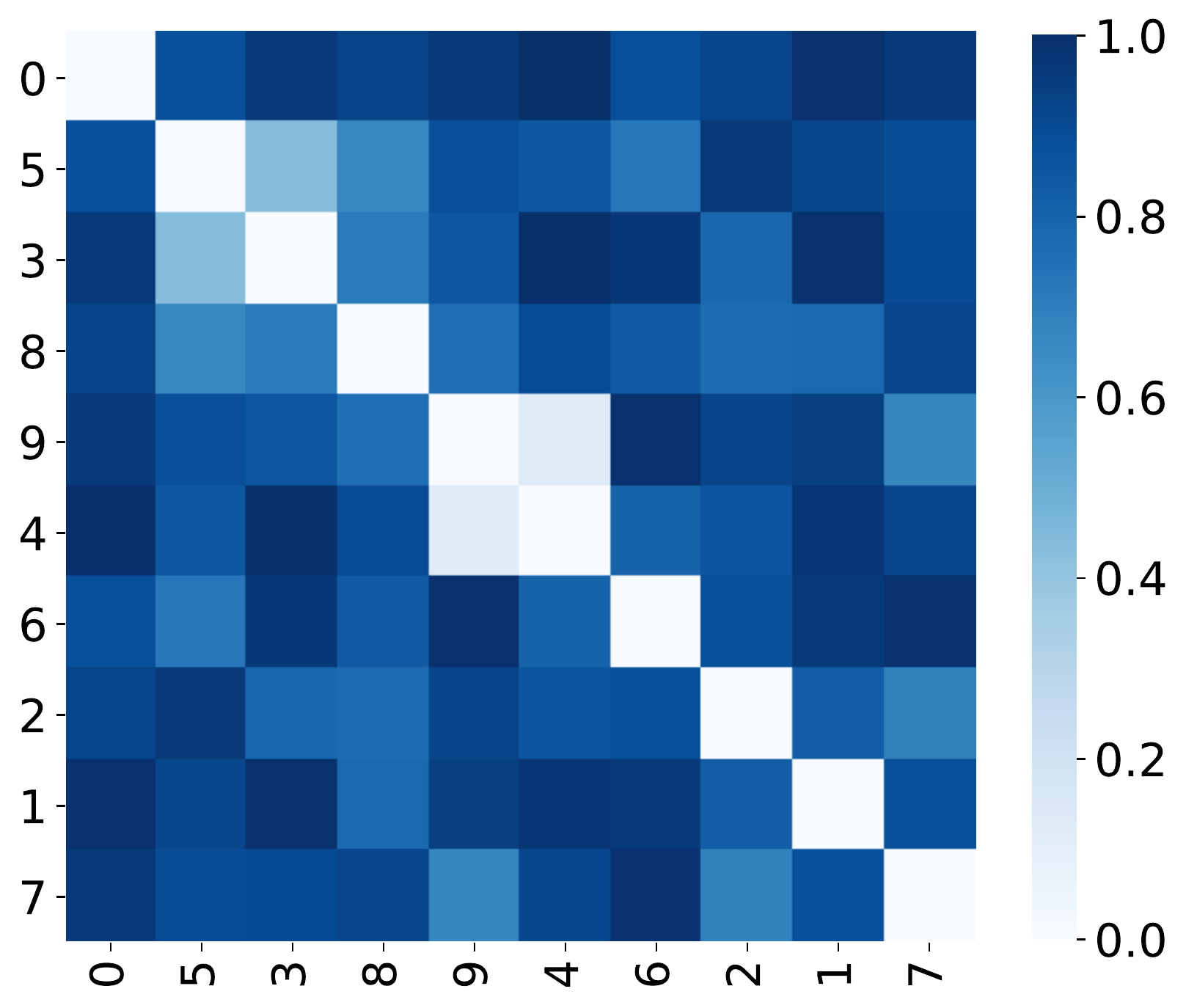}
        \caption{MNIST Visual Metric}
        \label{fig:metric_mnist}
    \end{subfigure}
    \begin{subfigure}[t]{0.31\textwidth}
        \centering
        \includegraphics[width=\textwidth]{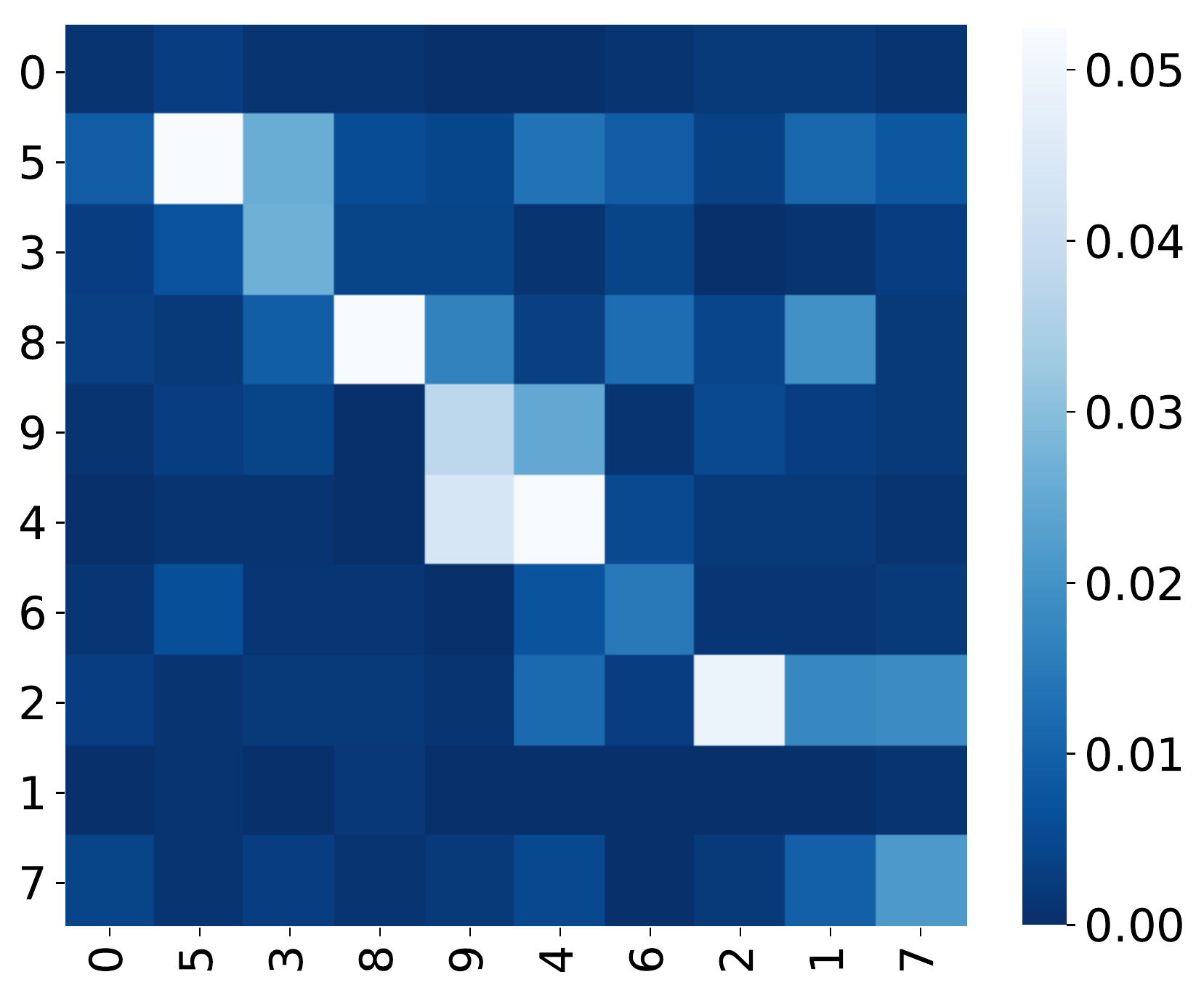}
        \caption{$\e = 20$}
        \label{fig:accuracy_gap_linear_20}
    \end{subfigure}
    \begin{subfigure}[t]{0.31\textwidth}
        \centering
        \includegraphics[width=\textwidth]{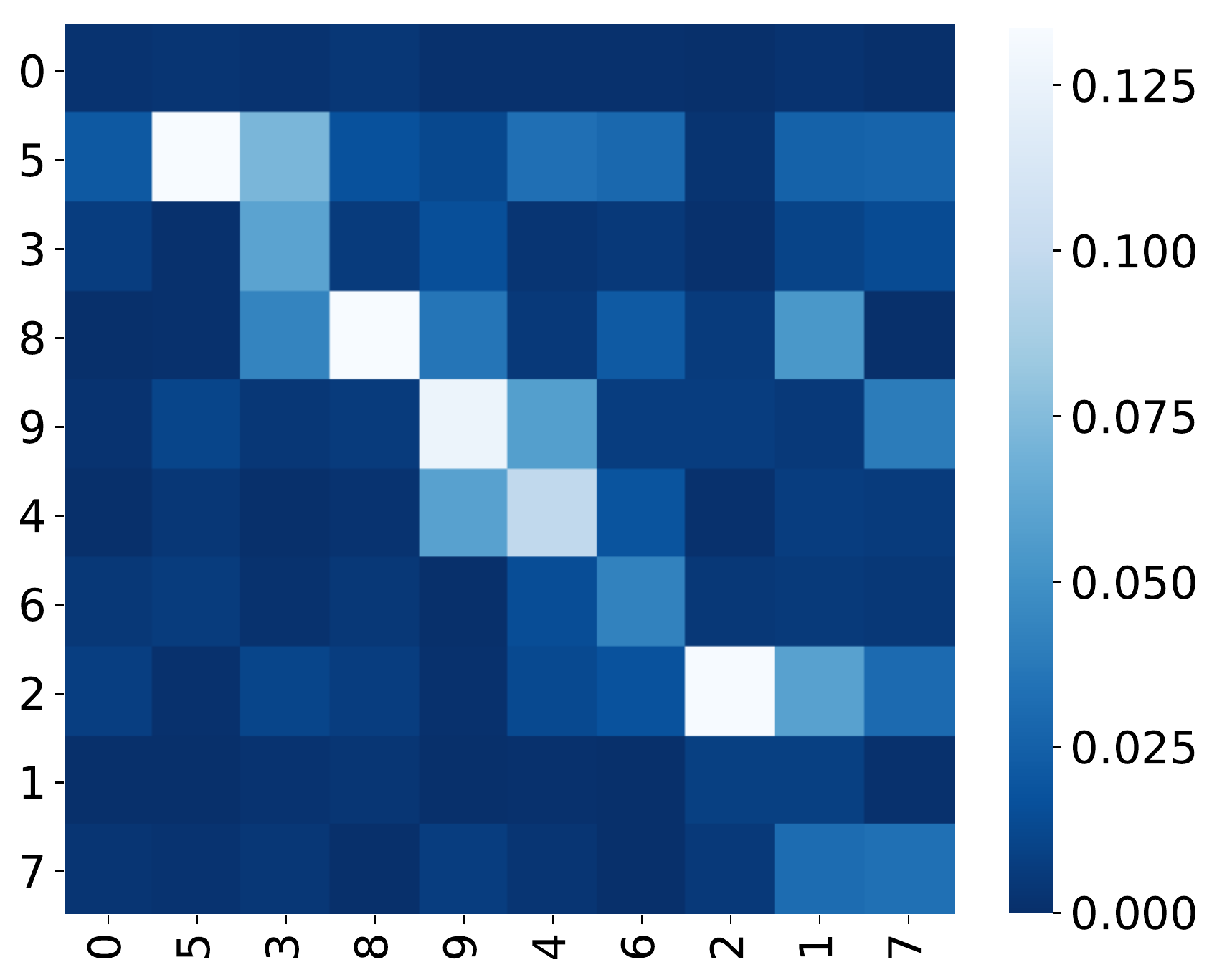}
        \caption{$\e = 38$}
        \label{fig:accuracy_gap_linear_38}
    \end{subfigure}
\caption{MNIST dataset: Accuracy gap $G$ between baseline model and PGD-trained model with $\e = 20$~\cref{fig:accuracy_gap_linear_20} and $\e = 38$~\cref{fig:accuracy_gap_linear_38}. The gap in accuracy caused by PGD training correlates with the visual metric. This causes the network to be less effective in fine-grained classification.}
\label{fig:accuracy_gap_mnist}
\end{figure*}
\begin{figure*}[htpb]
\centering
\captionsetup[subfigure]{justification=centering}
\begin{subfigure}[t]{0.29\textwidth}
        \centering
        \includegraphics[width=\textwidth]{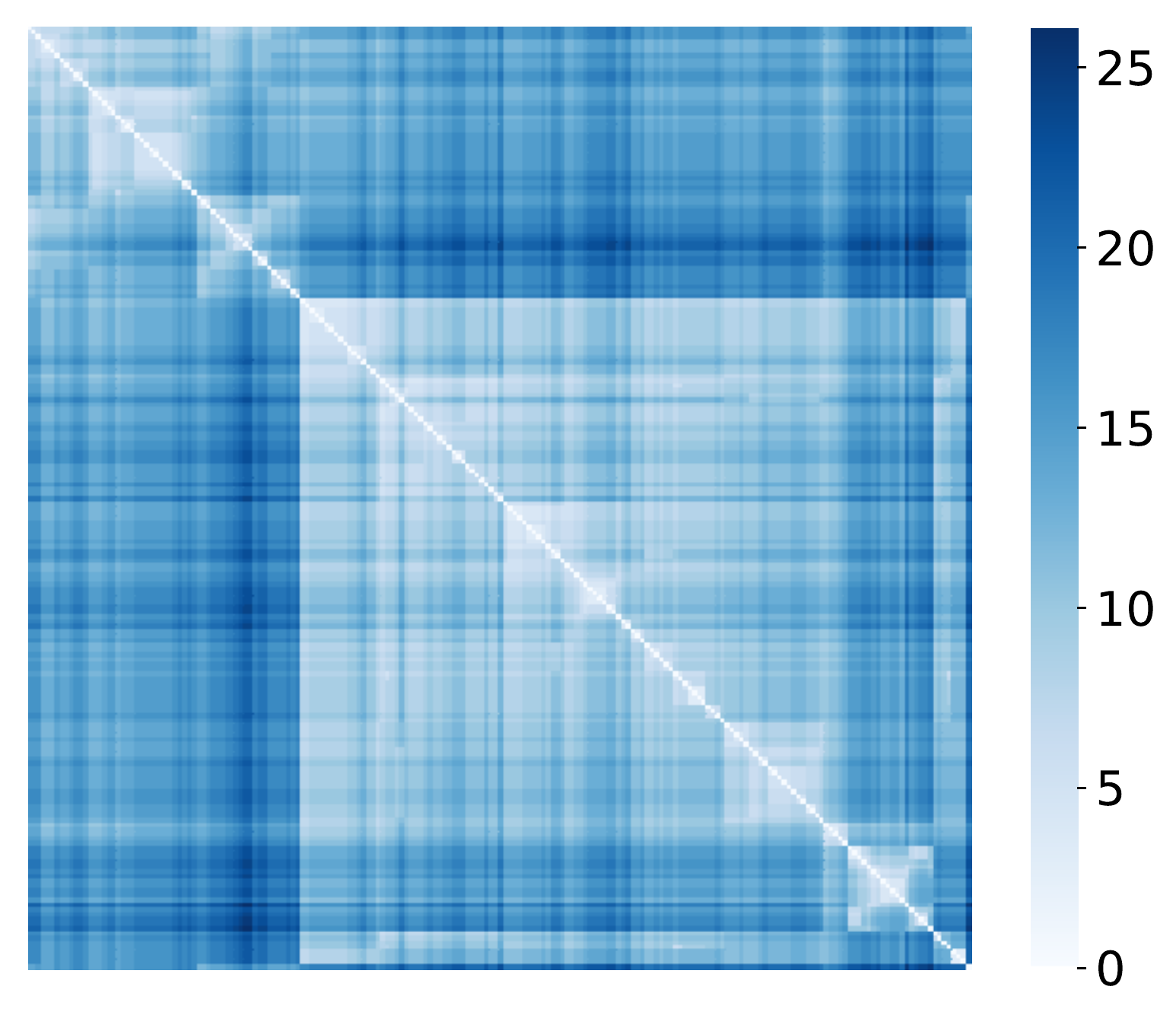}
        \caption{WordNet  \Ti}
        \label{fig:sm_tiny}
    \end{subfigure}
    \begin{subfigure}[t]{0.32\textwidth}
        \centering
        \includegraphics[width=\textwidth]{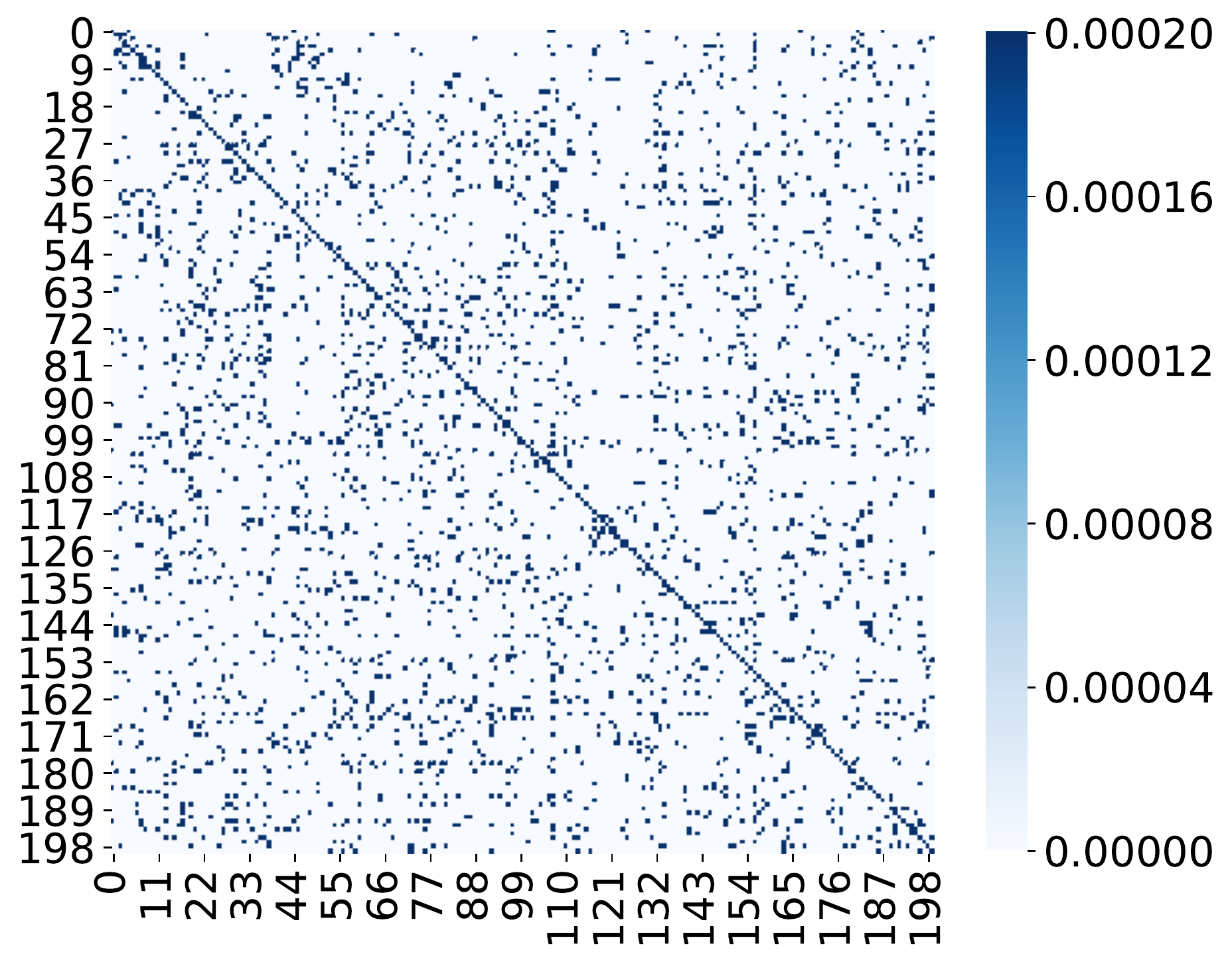}
        \caption{Accuracy gap}
    \end{subfigure}
    \begin{subfigure}[t]{0.36\textwidth}
        \centering
        \includegraphics[width=\textwidth]{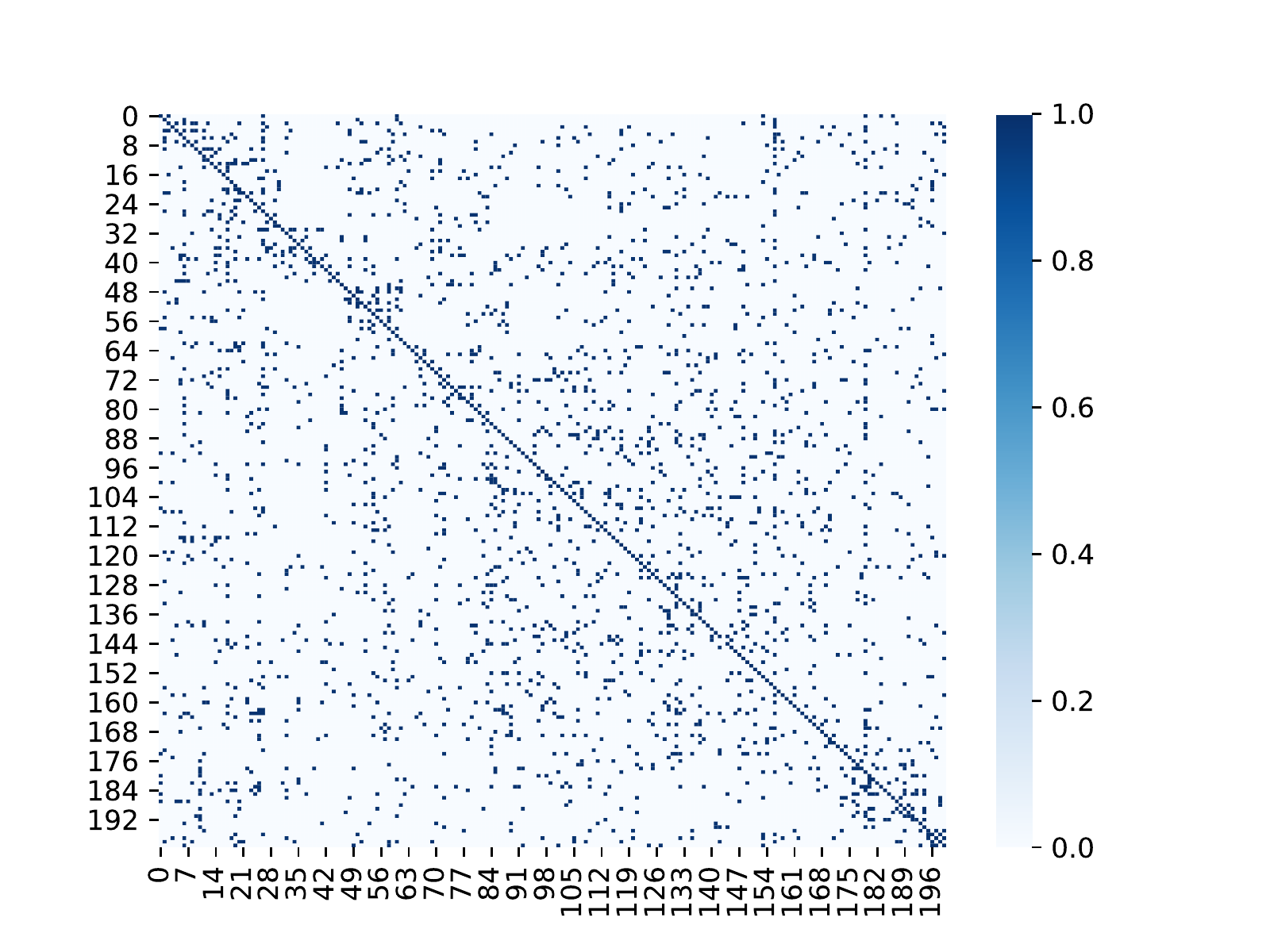}
        \caption{Confusion matrix of robust network.}
        \label{fig:metric_tiny}
    \end{subfigure}
\caption{Accuracy gap between baseline model and PGD-trained model with $\e = 8$ for \Ti.}
\label{fig:accuracy_gap_tiny}
\end{figure*}
\begin{figure*}[htpb]
\centering
\captionsetup[subfigure]{justification=centering}
\begin{subfigure}[t]{0.3\textwidth}
        \centering
        \includegraphics[width=\textwidth]{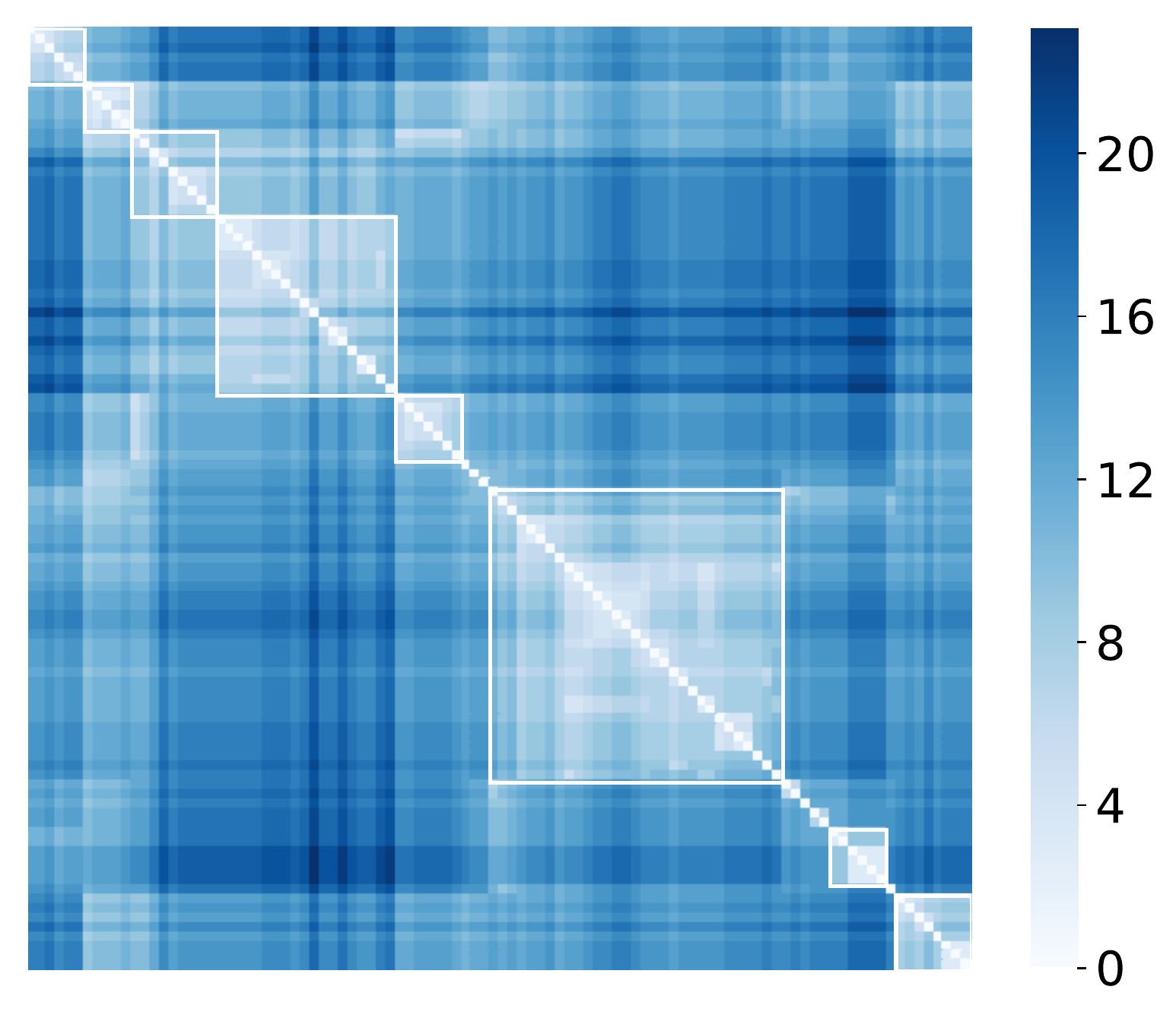}
        \caption{WordNet \Ch Metric}
        \label{fig:conf_eps8_c100}
    \end{subfigure}
    \begin{subfigure}[t]{0.32\textwidth}
        \centering
        \includegraphics[width=\textwidth]{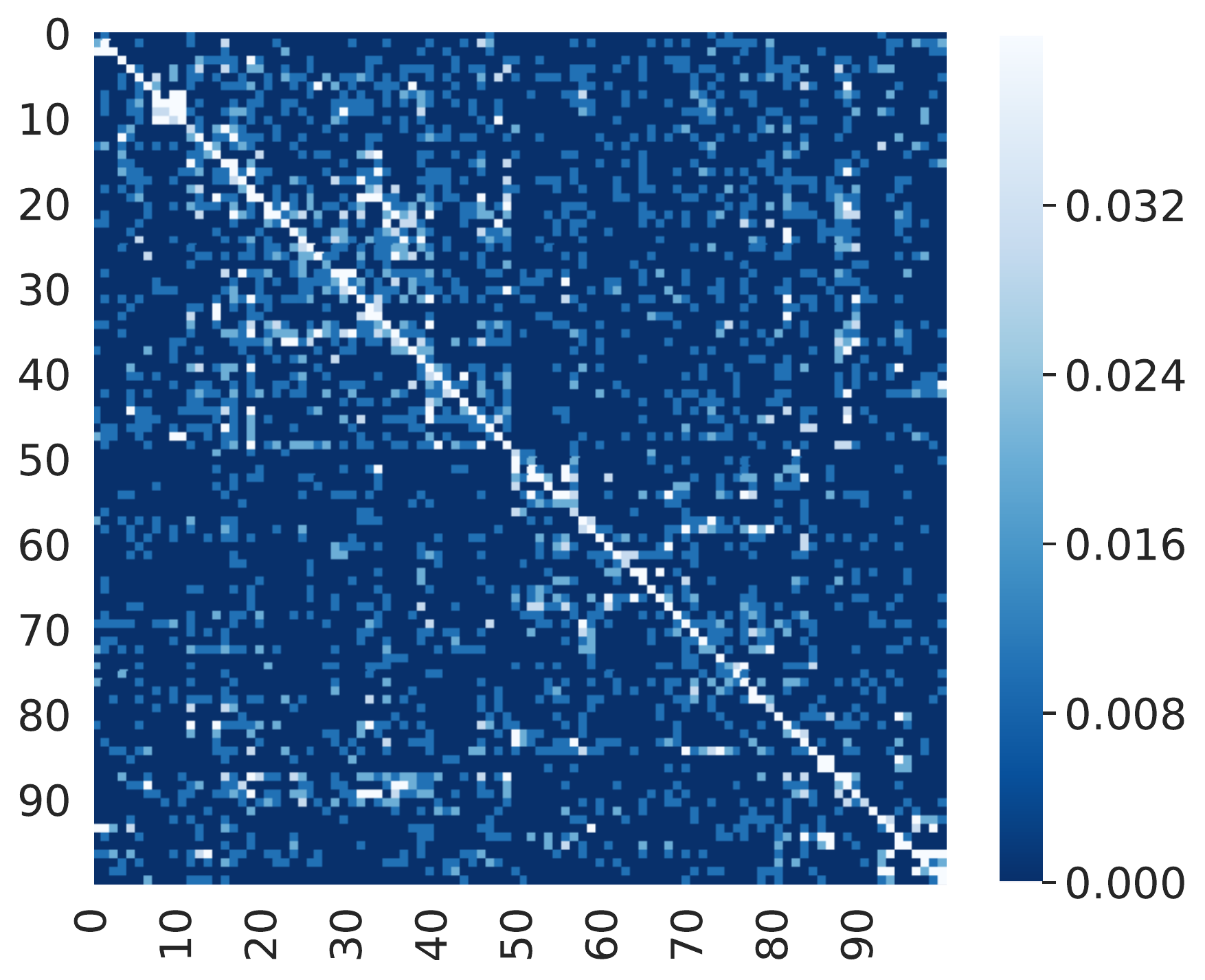}
        \caption{\allcnnl}
        \label{fig:accuracy_gap_cifar100_allcnnl_8}
    \end{subfigure}
    \begin{subfigure}[t]{0.32\textwidth}
        \centering
        \includegraphics[width=\textwidth]{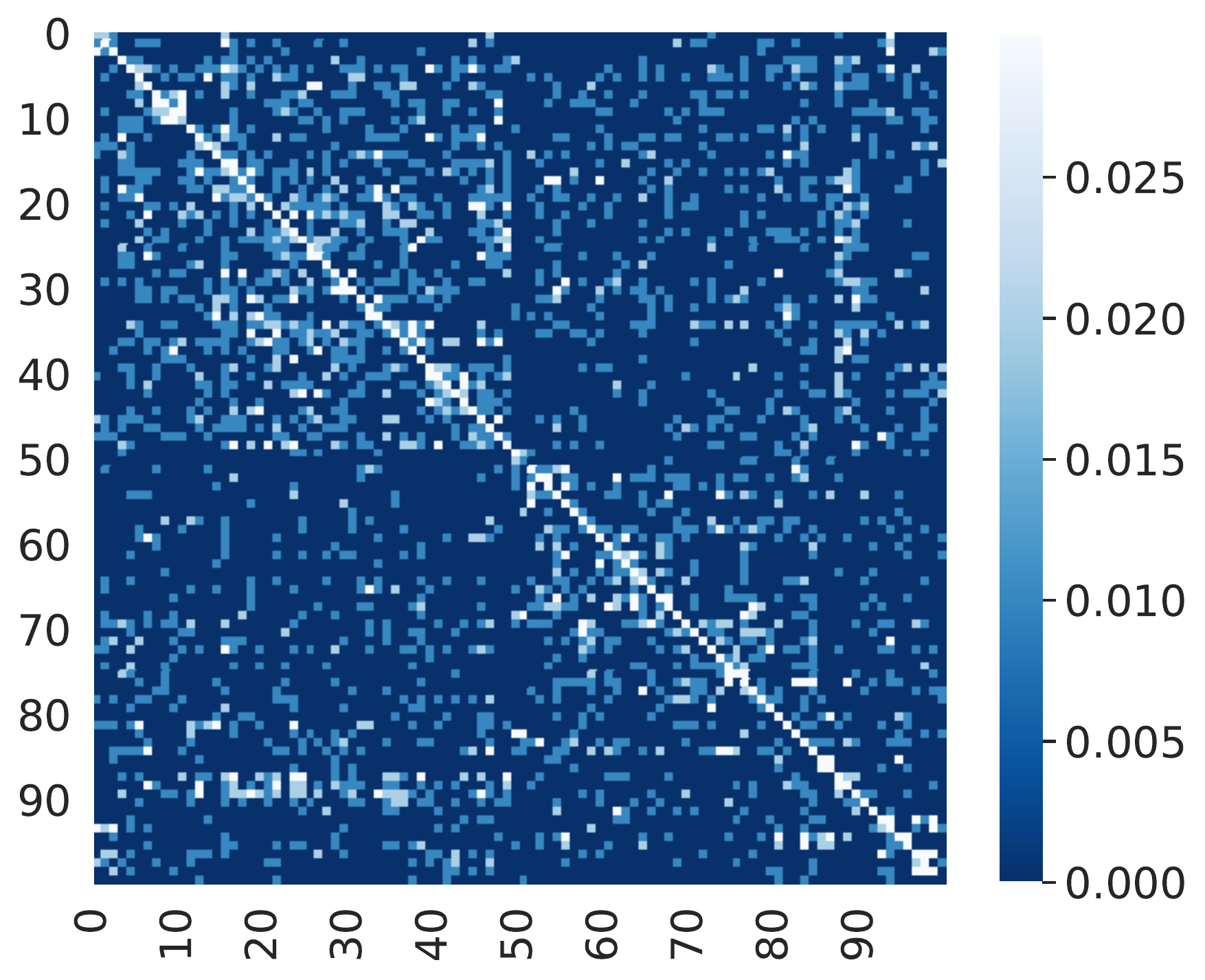}
        \caption{\wrnte}
        \label{fig:accuracy_gap_cifar100_wrn_8}
    \end{subfigure}
\caption{Accuracy gap between baseline model and PGD-trained model with $\e = 8$ for \Ch. The main cluster in the upper-left part of~\cref{fig:accuracy_gap_cifar100_allcnnl_8}  identifies animals in general. 
}
\label{fig:accuracy_gap_c100}
\end{figure*}

\cref{fig:c10_accuracy_gap}
 illustrates results for \Ct.
 In particular,~\cref{fig:accuracy_gap_wrn_8} shows the accuracy gap between a Wide Residual Network \cite{zagoruyko2016wide} trained using PGD and one trained with the standard cross-entropy loss. 
 From this figure, it is easy to see a visual correlation between metric and accuracy gap.
 Interestingly, the errors that are explained by such metrics, correspond to classes which are visually similar.
 For instance,~\cref{fig:c10_accuracy_gap} shows a gap on the pair \emph{bird-airplane} which are visually similar but semantically different. 
 Analogously, in~\cref{fig:accuracy_gap_mnist}, ~\cref{fig:accuracy_gap_tiny} and ~\cref{fig:accuracy_gap_c100} it is shown the WordNet metrics and the relative accuracy gaps for MNIST, Tiny-Imagenet and \Ch, respectively.
 Similar results are identifiable also for these datasets. In fact, regarding MNIST, not surprisingly, digits "$0$" and "$1$" hardly fool each other. 
 The most similar digits are "$4$" and "$9$": in fact, a small manipulation of such digits can be sufficient to make them indistinguishable.
 Also, regarding \Ch, as an example, from indices 8-11 there is an evident cluster composed by the classes \emph{man}, \emph{boy}, \emph{woman} and \emph{girl}. Other very connected classes are \emph{bridge}, \emph{skyscraper}, \emph{house}, \emph{castle} and \emph{road}. Moreover, there are animals that are semantically different but which are visually similar, such the couple 32-90 that are \emph{seal} and \emph{otter} respectively. The bottom-right cluster represents flowers and plants.
 
In~\cref{tab:gap_coff} a quantitative measure (supporting the aforementioned 'visual' results) of the correlation between accuracy gap and relative metric is provided. The minus sign is due to the fact that confusion matrices and distances are inversely correlated: when the values of diagonal increase of the confusion matrices, then the distance between classes decreases, on average. 
For MNIST the correlation is higher since an approximation of the actual visual metric has been used, while for \Ct~and \Ch~the correlation is lower because some pairs, for example, bird-airplane are semantically different. Moreover, it is remarked that with high output dimension, the correlation decreases even when there are well-correlated structures. The correlation between two random matrices in $\bR^{200\times 200}$ is almost zero in expectation.

{
\footnotesize
\begin{table}[!htp]
\centering
\begin{tabularx}{\columnwidth}{p{2cm}X X X X X}
\toprule
 &\tbf{MNIST} & \tbf{CIFAR-10} & \tbf{CIFAR-100} & \tbf{Tiny-Imagenet}\\
\cmidrule{2-5}
Correlation &-0.88 & - 0.65 & -0.35 & -0.22\\
\bottomrule
\end{tabularx}
\caption{Correlation $\rho$ between accuracy gap and relative metric for all the datasets.}
\label{tab:gap_coff}
\end{table}
}

Given these premises and observations, the following conjecture can be made: when the number of classes is high than boundaries among similar classes becomes more complex. Thus, as an ablation study, two 2-classes problems with the \Ct dataset are reported in the following: the first problem is to distinguish classes \emph{airplane} (id: 0) and \emph{horse} (id: 7) while the second is \emph{cat} (id: 3) vs \emph{dog} (id: 5). In ~\cref{fig:two_class} it is shown that even in simple settings, adversarial training affects dramatically fine-grained classification. 

\subsection{Unbalanced classification}
\label{ss:unbalance}
Although real-world datasets are long-tailed~\cite{deng2009imagenet}, most of the experiments and theoretical findings on the accuracy-robustness trade-off in the literature were performed with balanced datasets~\cite{tsipras2018robustness}. 

Through an experimental analysis, it is shown that when classes are unbalanced, adversarial training can have dramatic effects on clean accuracy. 
For this analysis, the same 2-classes problems of the ablation study reported in subsection \ref{ss:effects}
 are selected: cat-vs-dog and airplane-vs-horse. Classes are artificially randomly unbalanced such that their ratio is 0.3.

The cat-vs-dog classification problem is intrinsically difficult since the two classes have many features in common. 
Moreover, \Ct have low-resolution images making (sometimes) this classification task not trivial also for human classifiers.
On the contrary, airplane-vs-horse is a simple task and thus one should expect that adversarial training does not decrease much clean accuracy.

\begin{figure*}[htpb]
\centering
    \begin{subfigure}[t]{0.48\textwidth}
        \centering
        \includegraphics[width=\textwidth]{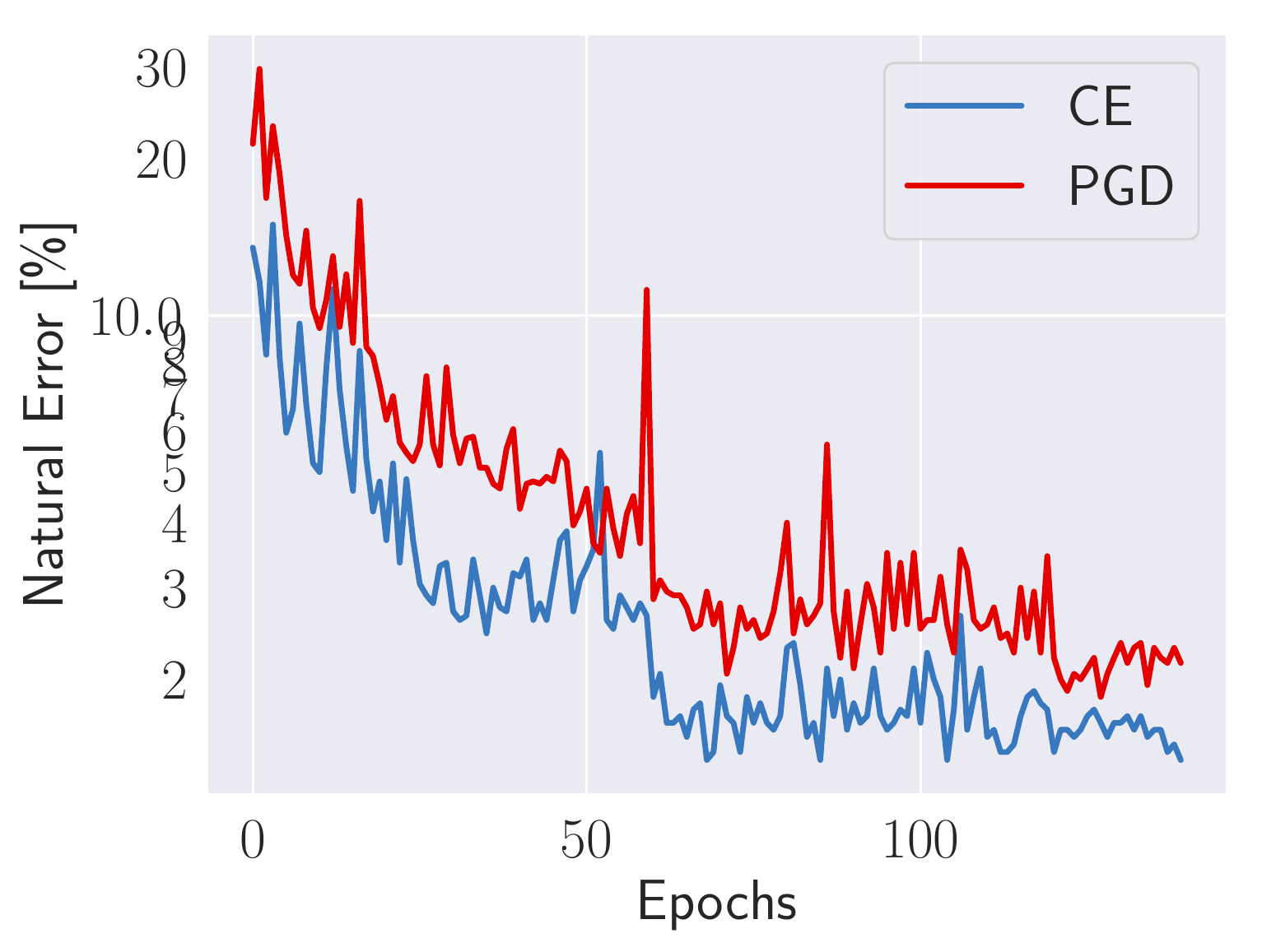}
         \caption{Standard training 0-vs-7.}
        \label{fig:07nat}
    \end{subfigure}
    \begin{subfigure}[t]{0.48\textwidth}
        \centering
        \includegraphics[width=\textwidth]{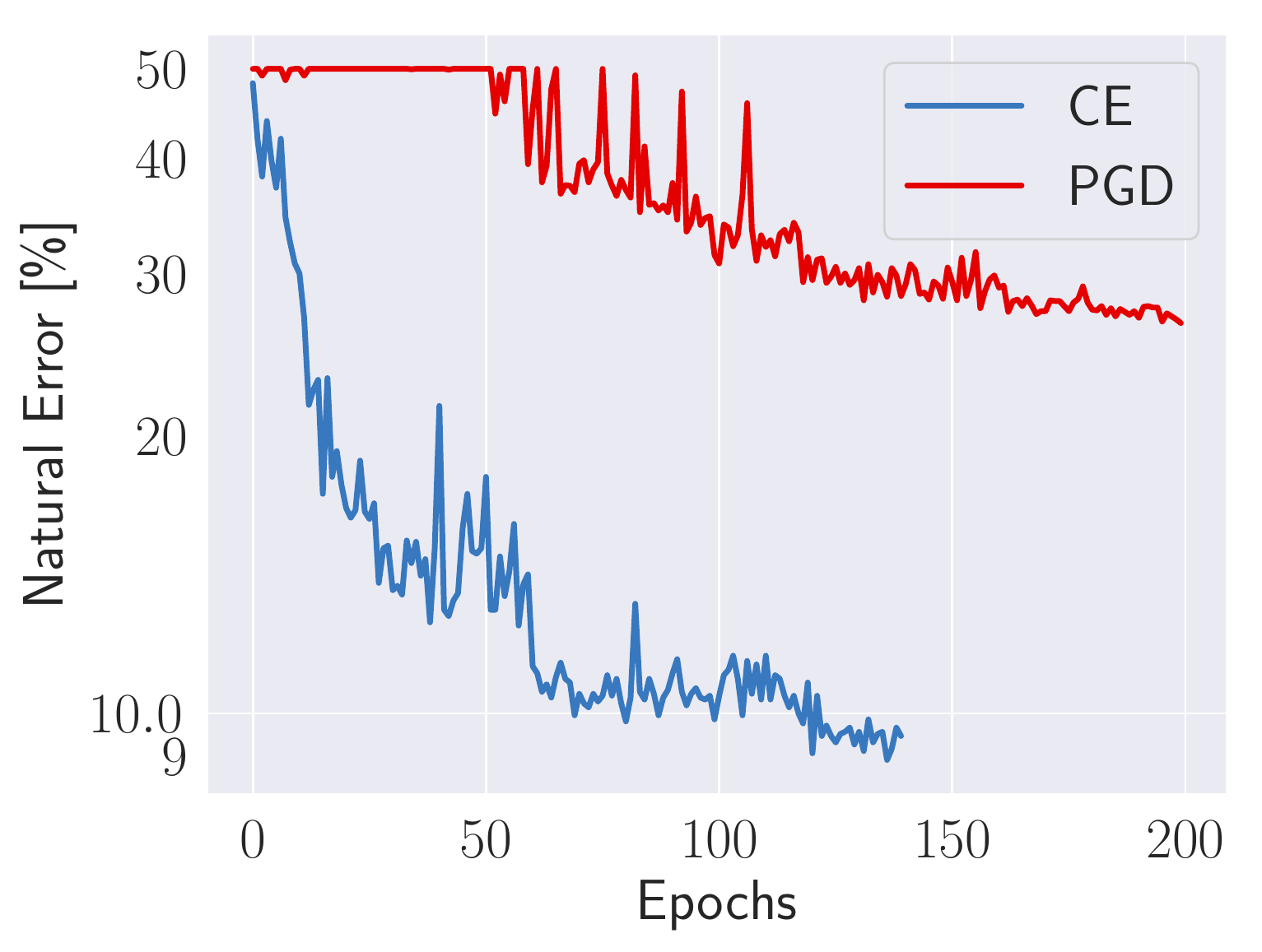}
        \caption{Standard training 3-vs-5.}
        \label{fig:35nat}
    \end{subfigure}\\
    \begin{subfigure}[t]{0.48\textwidth}
        \centering
        \includegraphics[width=\textwidth]{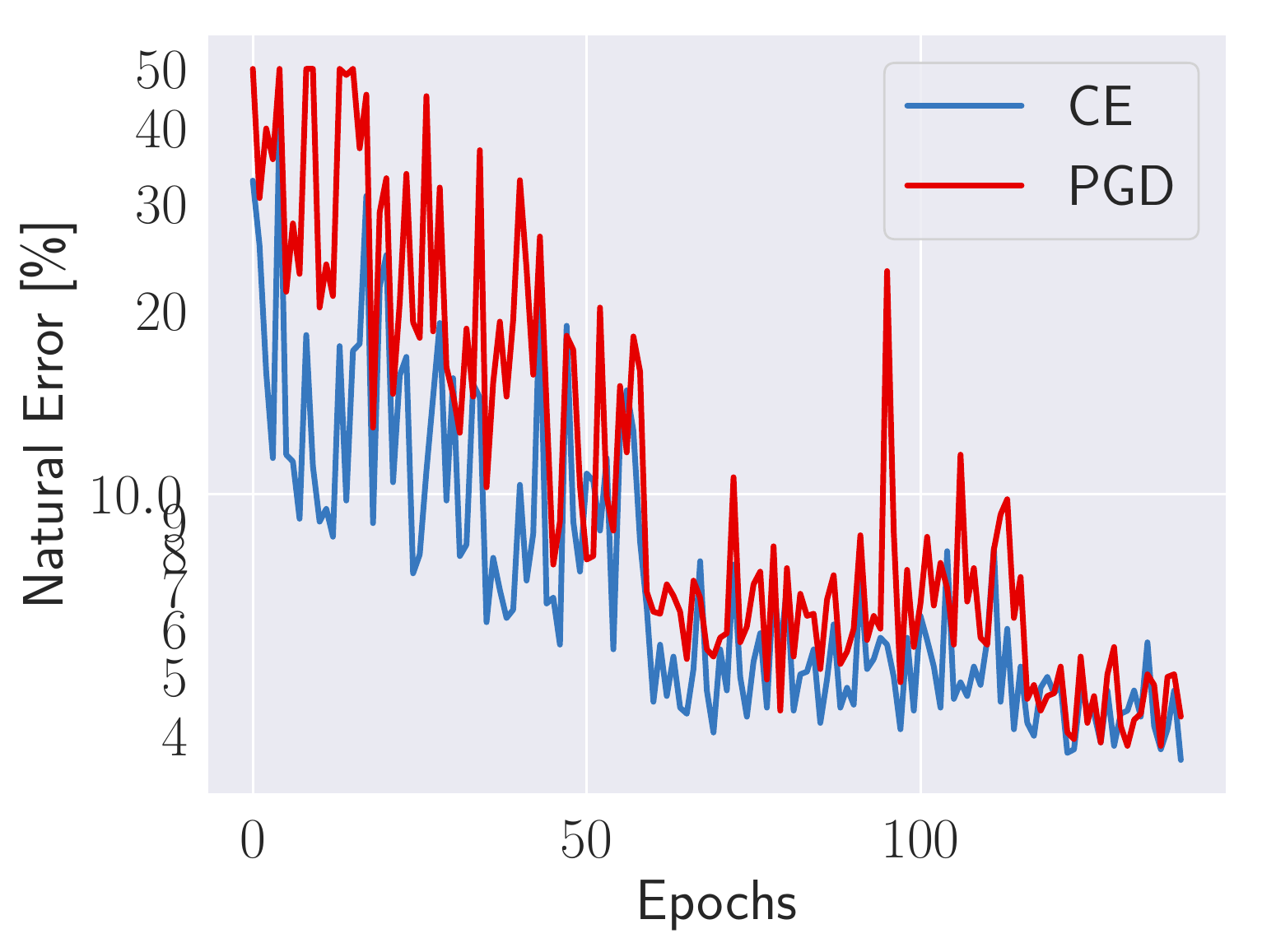}
        \caption{Unbalanced 0-vs-7.}
        \label{fig:07nat_un}
    \end{subfigure}
    \begin{subfigure}[t]{0.48\textwidth}
        \centering
        \includegraphics[width=\textwidth]{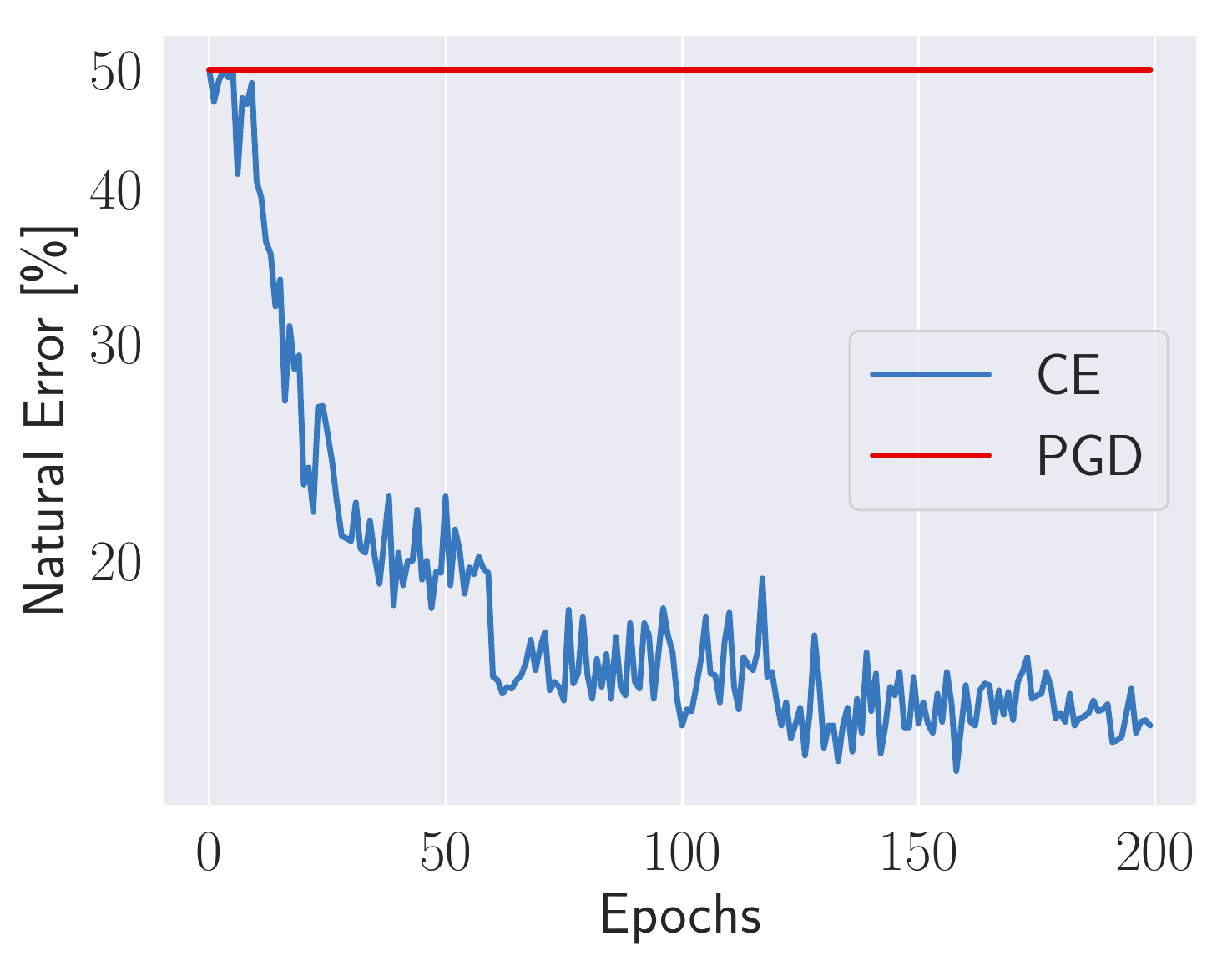}
        \caption{Unbalanced 3-vs-5.}
        \label{fig:35nat_un}
    \end{subfigure}
 \caption{Panel (a) and Panel (b) report validation curves for the problems 0-7 and 3-5, respectively. Panel (c) and Panel (d) are the same curves with a randomly unbalanced dataset. For classes (3) and (5), when the dataset is unbalanced the validation error remains constant to random choice (50\%).}
 \label{fig:two_class}
\end{figure*}

The results of these two experiments are shown in~\cref{fig:two_class}: two different considerations are here reported. The first is that when classes are similar, as mentioned above, PGD heavily impacts on the performance with respect to standard training. Instead, for dissimilar classes, the effect is much less pronounced. This a solid argument for supposing that using a single $\e$ may be not optimal.
The second consideration is that when dataset is unbalanced, PGD further amplifies the difficulty of the classification task. For example, for cat-vs-dog (\cref{fig:35nat_un}), in presence of unbalance, the model can't be fit at all.

\subsection{Entropy of softmax outputs}
\label{subsec:entropy_reg} 
\begin{figure*}
    \centering
    \includegraphics[width=0.45\textwidth]{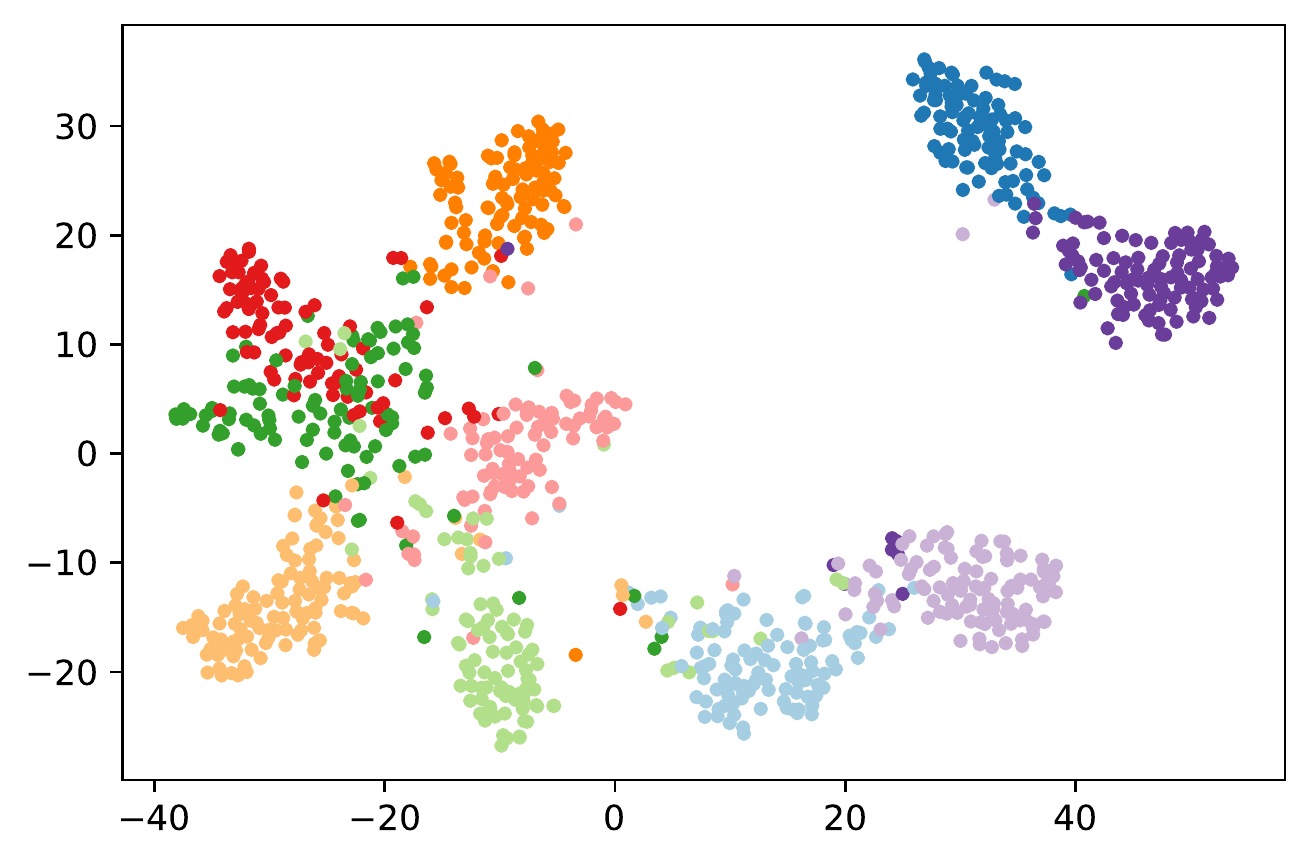}
    \includegraphics[width=0.45\textwidth]{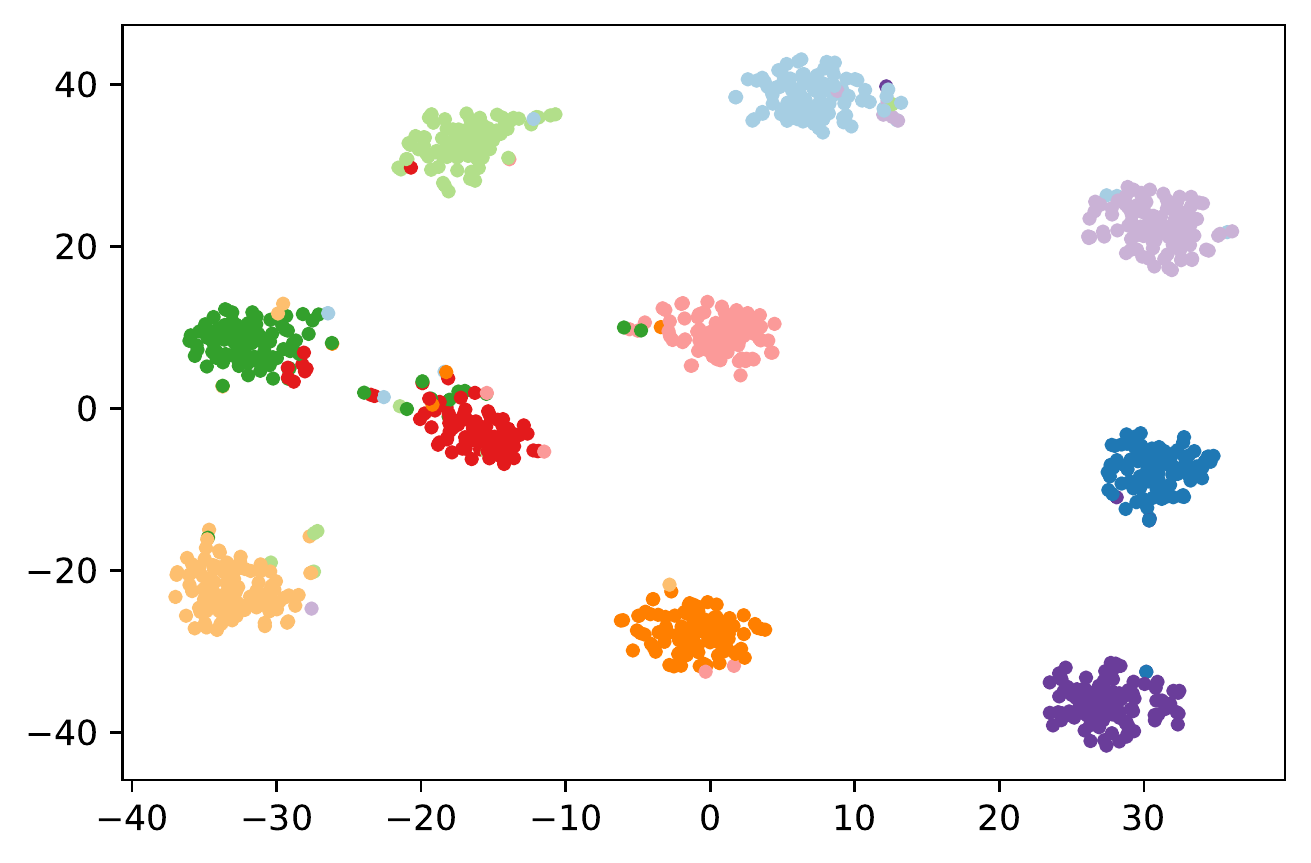}
    \caption{\Ct. Comparison of t-SNE computed on the representation of standard models (left) and robust models (right). Each coloured cluster correspond to one particular class. }
    \label{fig:tsne}
\end{figure*}

\begin{figure*}[htpb]
\centering
\captionsetup[subfigure]{justification=centering}
    \begin{subfigure}[t]{0.45\textwidth}
        \centering
        \includegraphics[width=\textwidth]{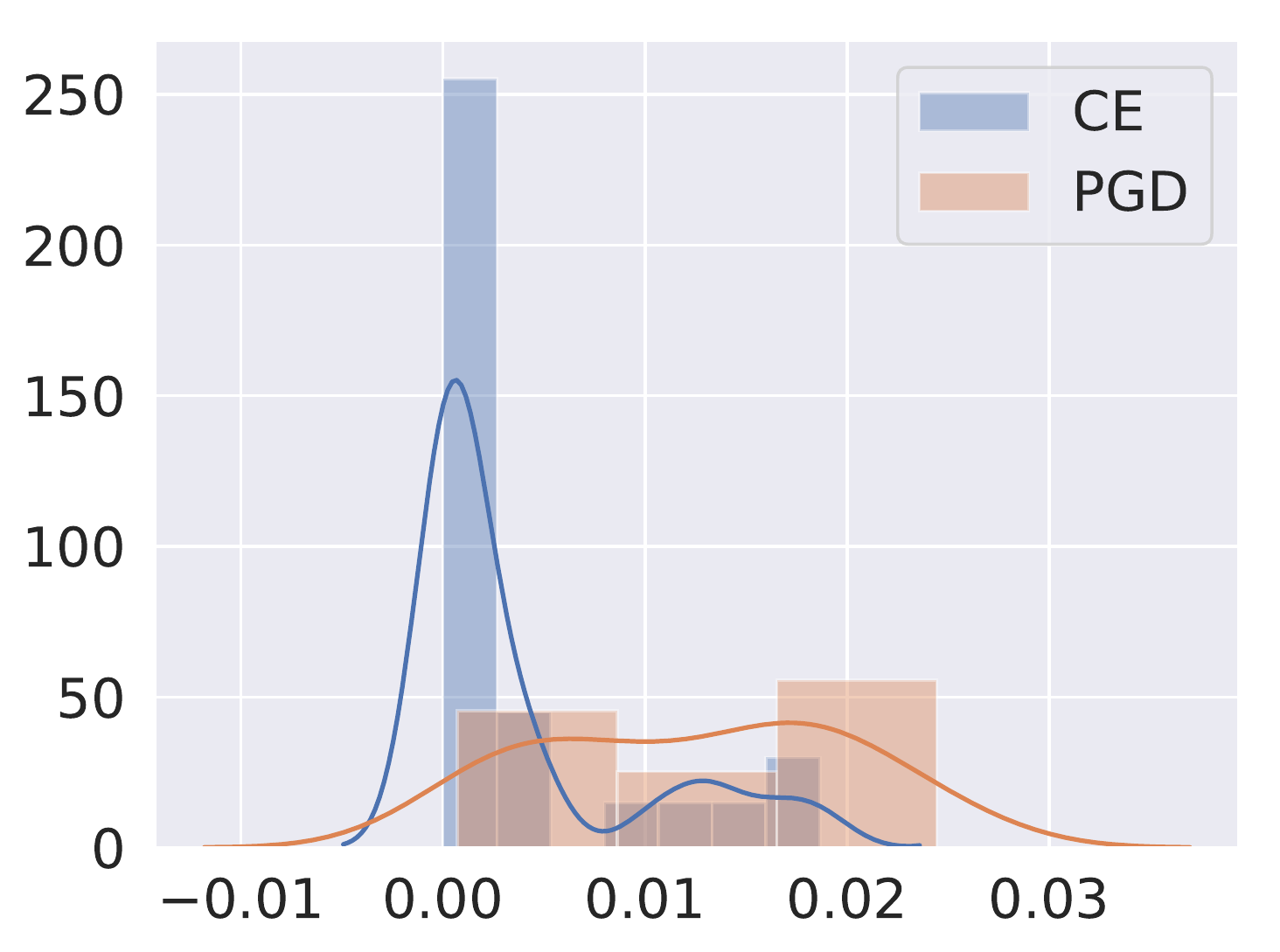}
        \caption{\Ti}
        \label{fig:conf_eps8_tiny}
    \end{subfigure}
    \begin{subfigure}[t]{0.45\textwidth}
        \centering
        \includegraphics[width=\textwidth]{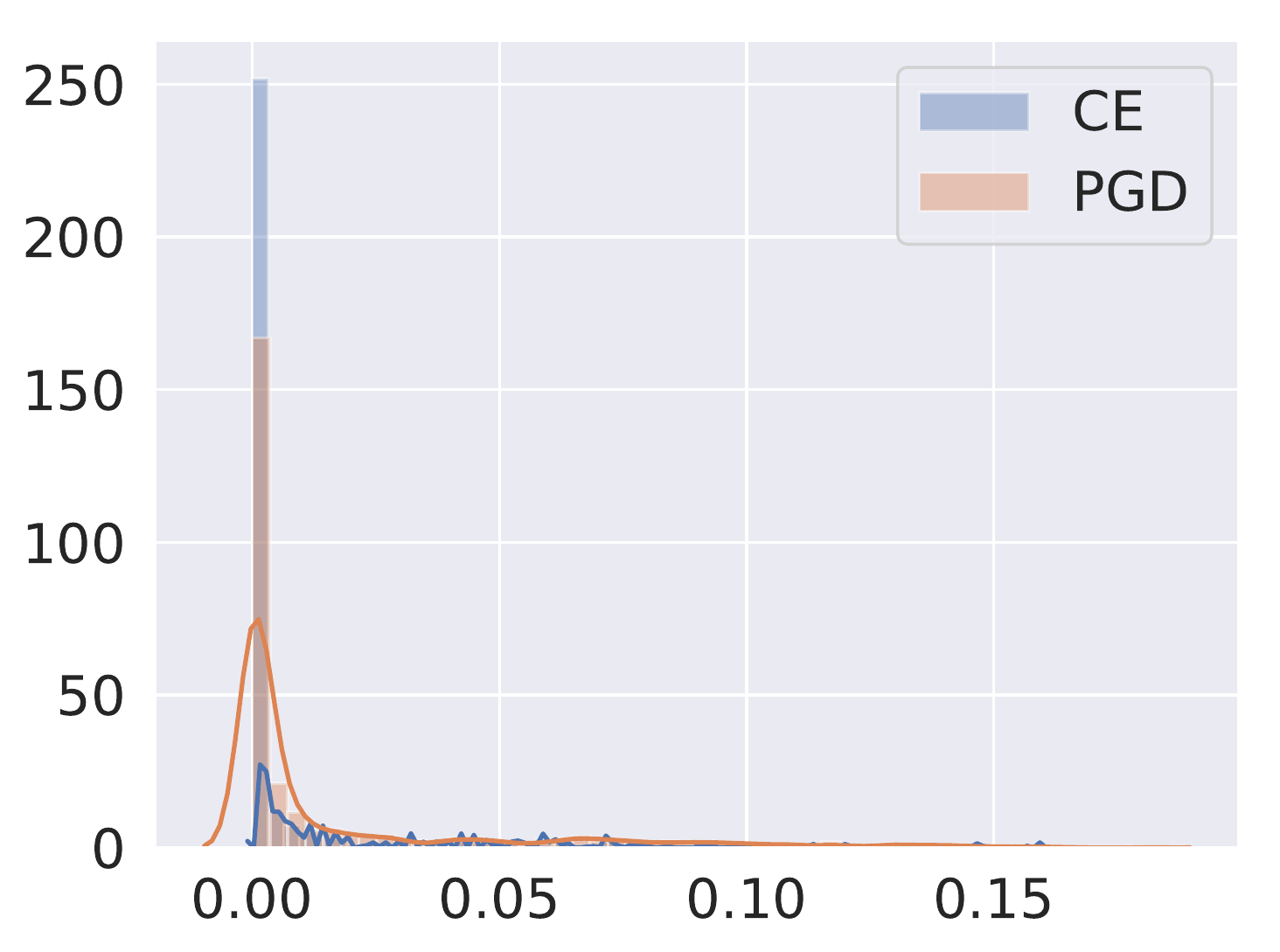}
        \caption{\Ct}
        \label{fig:conf_eps16}
    \end{subfigure}
\caption{Entropy histograms of prediction confidence for \wrnst \, with $\e = 8$ of class \emph{airplane}. Robust networks provide more conservative predictions. Adversarial training prevents the network to provide high confidence predictions. This is a consequence of simplifying boundaries as shown in ~\cref{ss:boundaries}. Other classes follow the same trend.}
\label{fig:confidence}
\end{figure*}

One of the issues of 'standardly' trained network, is that they are over-confident, that is, they tend to predict classes with with high probability even when images are not clear~\cite{guo2017calibration}.
Adversarial training can be seen as an implicit regularization and thus it is legitimate to analyze confidence of predictions on robust models. Indeed, in~\cref{fig:confidence} it is shown that another characteristic of adversarial training is reducing confidence of predictions; in fact, the entropy of class logits of the robustly trained network is much higher.
This suggests that confidence scores obtained by thresholding the softmax predictions should be changed. 
Thus, it may seems that robust representations are less discriminative than standard ones. 
\footnote{For those who are not used to deep learning language, in this context a representation is the vector (output of the feature extractor) that is feed to the last layer which is a linear classifier.}
It turns out that this intuition is true and supported by~\cref{fig:tsne}. 
In order to assess the structure of representations, it has been employed t-SNE~\cite{maaten2008visualizing}, a techniques that allows to visualize high-dimensional data in 2 or 3 dimensions. 
From~\cref{fig:tsne}, it is clear that robust representations are less clustered with respect to natural ones. Each coloured cluster correspond to one particular class. 


\subsection{PGD flattens boundaries}
\label{ss:boundaries}
In order to better understand the behavior of PGD and also to compare it with WPGD defined in~\cref{s:wpgd}, a simple classification problem with three classes is considered.
~\cref{fig:example_PGD} and~\cref{fig:example_WPGD} show the boundaries for PGD and WPGD (for different $\e$), respectively. ~\cref{fig:tr_class_orig} represents the standard training with which achieves almost zero error. As $\e$ increases boundaries are more flattened as orthogonal as possible to the gradient direction. The adopted cost matrix is $C=\bmat{0&10 & 0.01\\ 10 &0&1 \\ 0.01&1 & 0}$.
Related this results,~\cite{Moosavi-Dezfooli:2018aa} showed experimentally that the main effect of PGD is to reduce the curvature of boundaries. However, it can be easily shown that even when the curvature is, robust training still has an effect. Moreover, it is noticed that gradients are more aligned to the vector which connect two classes. This is due to the "isotropic" effect of PGD which tend to estimate more isotropic distributions. This is in accordance with \cite{tsipras2018robustness} in which authors observed that gradient on the robust model are more meaningful.
This argument is also in accordance to results on fine-grained classification present on this work, suggesting that visually similar are separated by more complex boundaries. Instead, WPGD controls the the regularization of boundaries through the cost matrix: boundaries for couple of classes considered more similar are mostly preserved. 

\newcommand{\sz}{0.49}
\begin{figure*}[htpb]
	\begin{subfigure}[t]{\sz\textwidth}
	\centering
	\includegraphics[width=\textwidth]{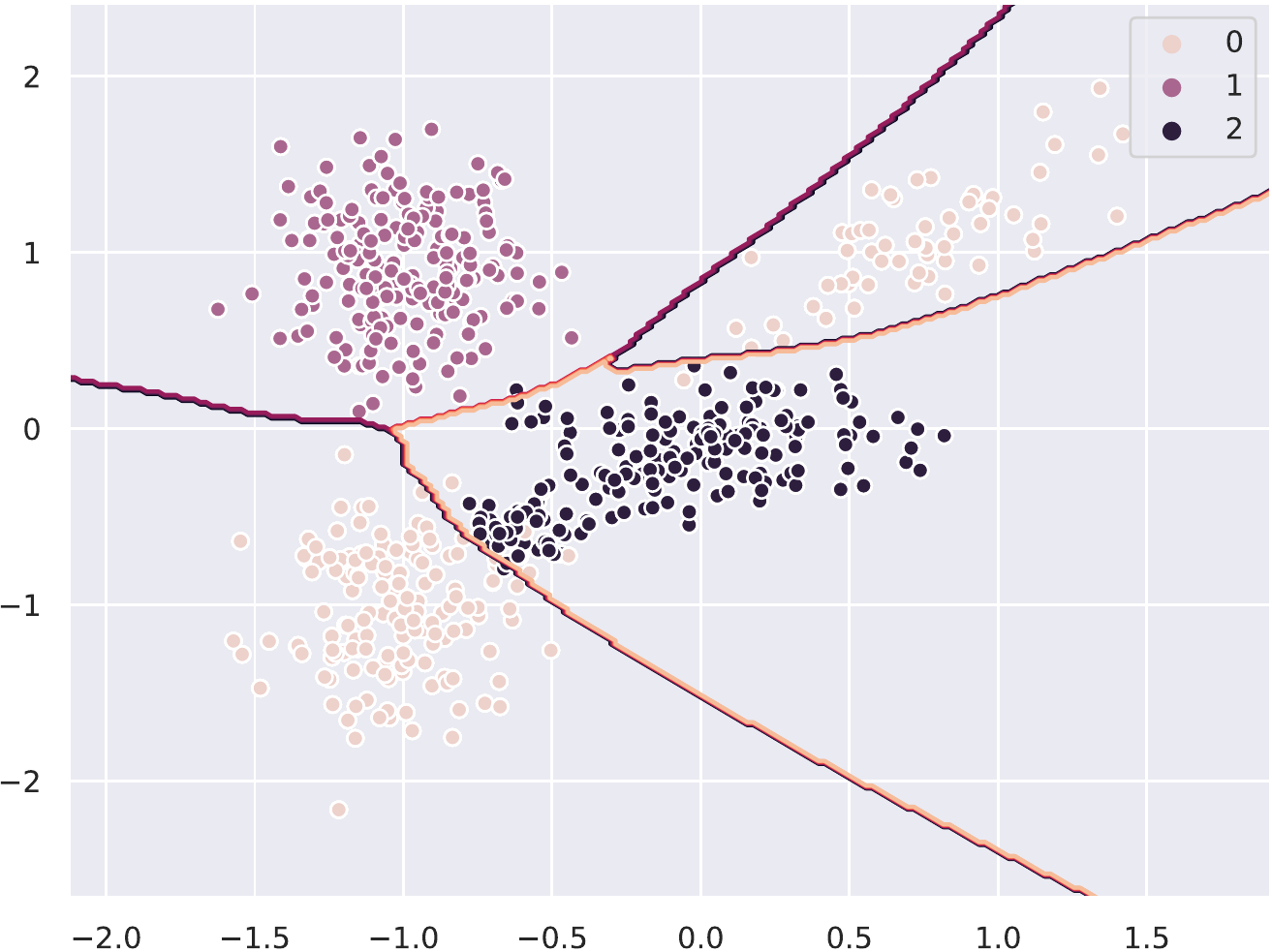}
	\caption{Original problem}
	\label{fig:tr_class_orig}
	\end{subfigure}
	\begin{subfigure}[t]{\sz\textwidth}
	\centering
	\includegraphics[width=\textwidth]{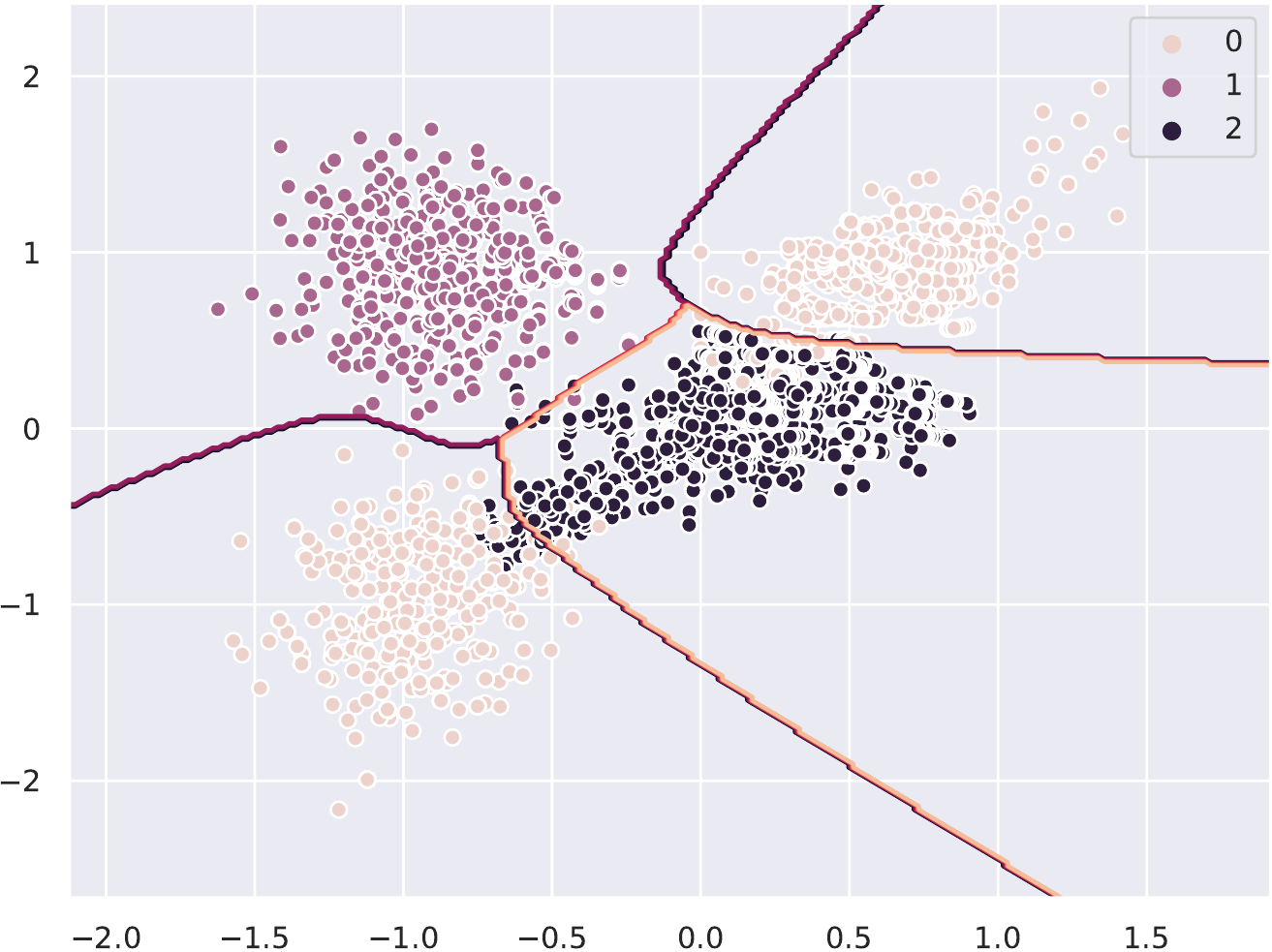}
	\caption{PGD ($\e = 0.2$)}
	\label{fig:tr_class_50}
	\end{subfigure}
	\begin{subfigure}[t]{\sz\textwidth}
	\centering
	\includegraphics[width=\textwidth]{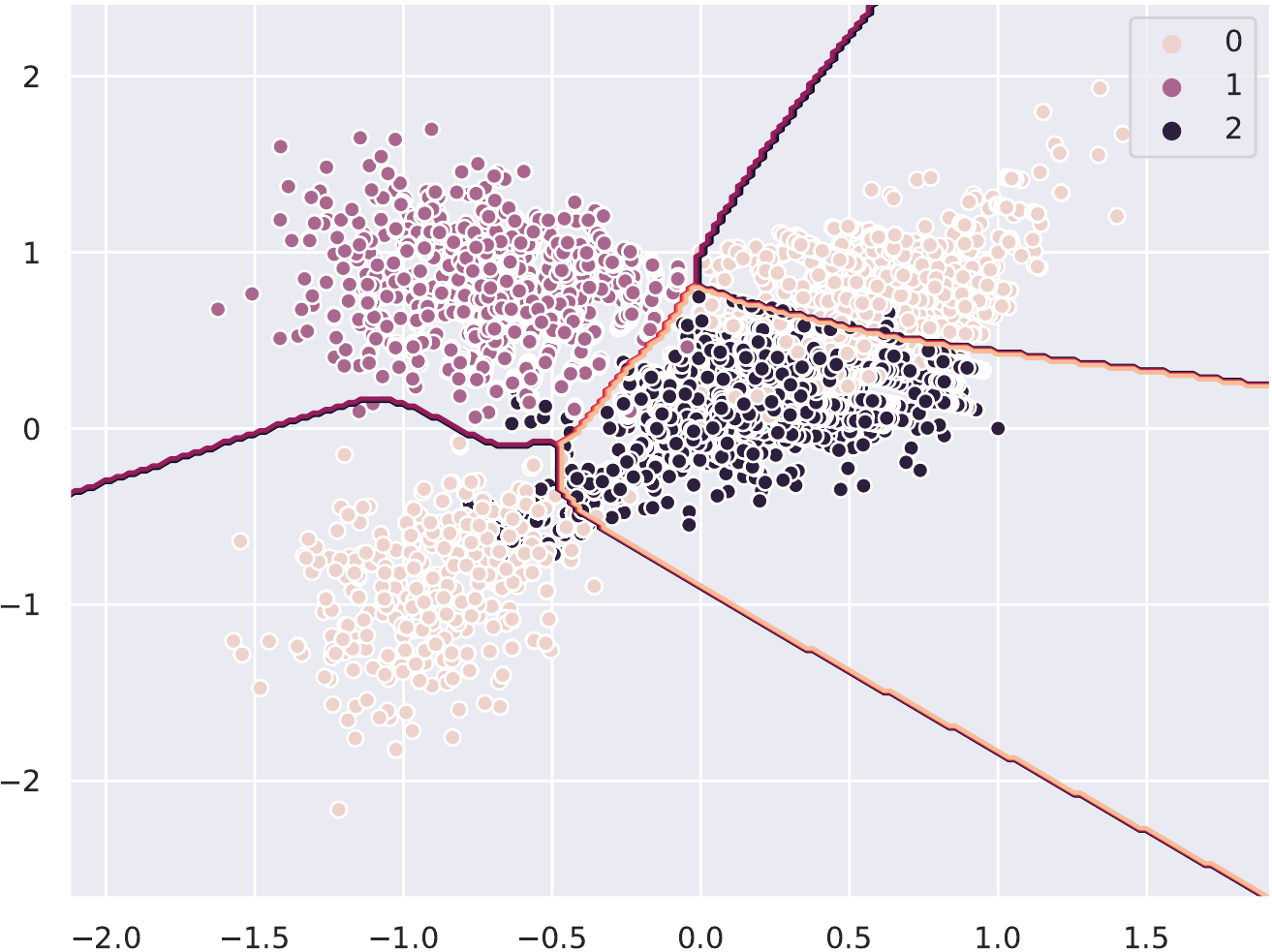}
	\caption{PGD ($\e = 0.4$)}
	\label{fig:tr_class_100}
	\end{subfigure}
	\begin{subfigure}[t]{\sz\textwidth}
	\centering
	\includegraphics[width=\textwidth]{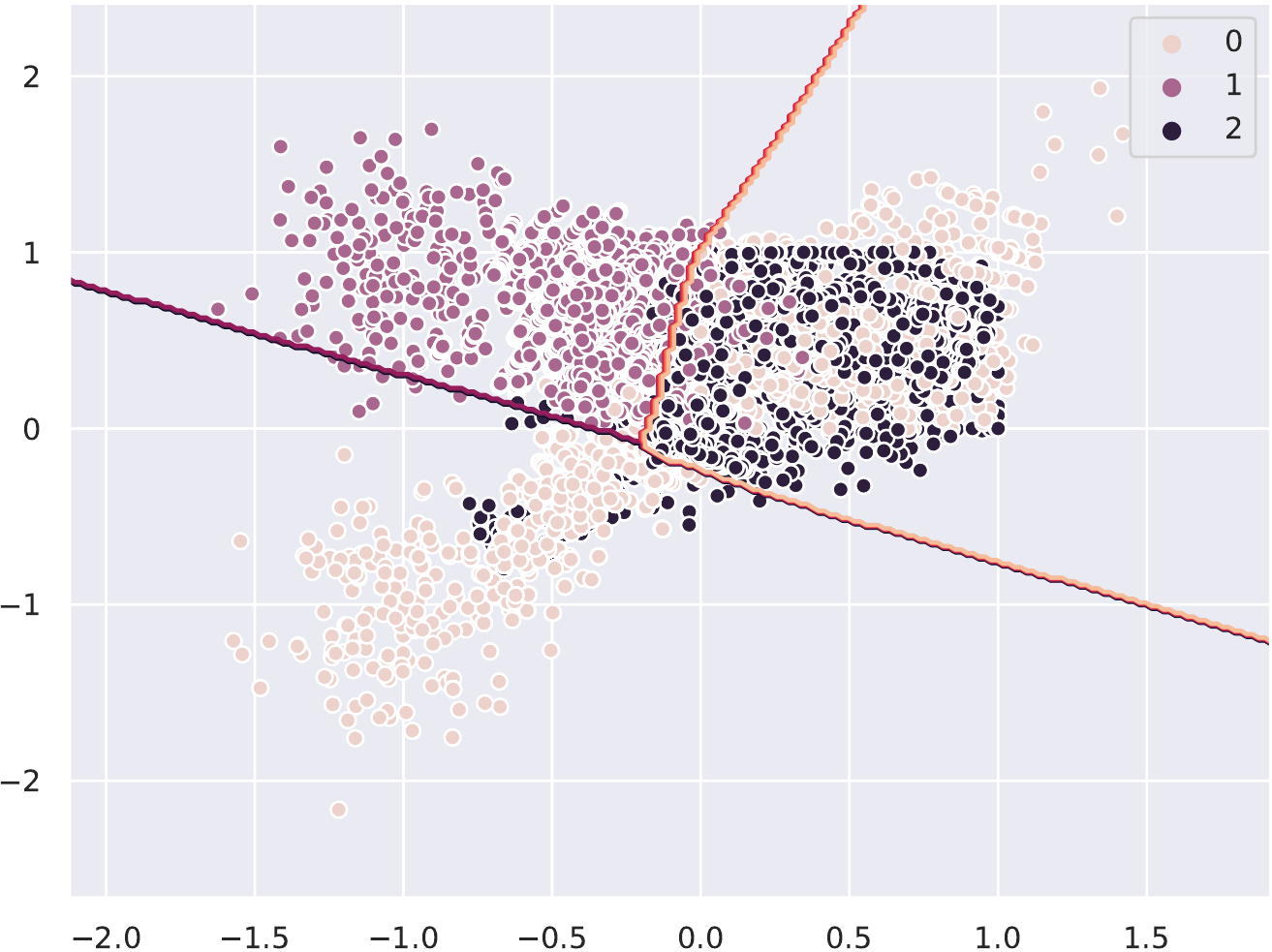}
	\caption{PGD ($\e = 0.8$)}
	\label{fig:tr_class_200}
	\end{subfigure}
\caption{Effect of robust $\ell_2$-training on a simple classification problem. PGD training flattens the boundaries.}
\label{fig:example_PGD}
\end{figure*}
\begin{figure*}[htpb]
	\begin{subfigure}[t]{\sz\textwidth}
	\centering
	\includegraphics[width=\textwidth]{example_0.pdf}
	\caption{Original problem}
	\label{fig:tr_class_orig_2}
	\end{subfigure}
	\begin{subfigure}[t]{\sz\textwidth}
	\centering
	\includegraphics[width=\textwidth]{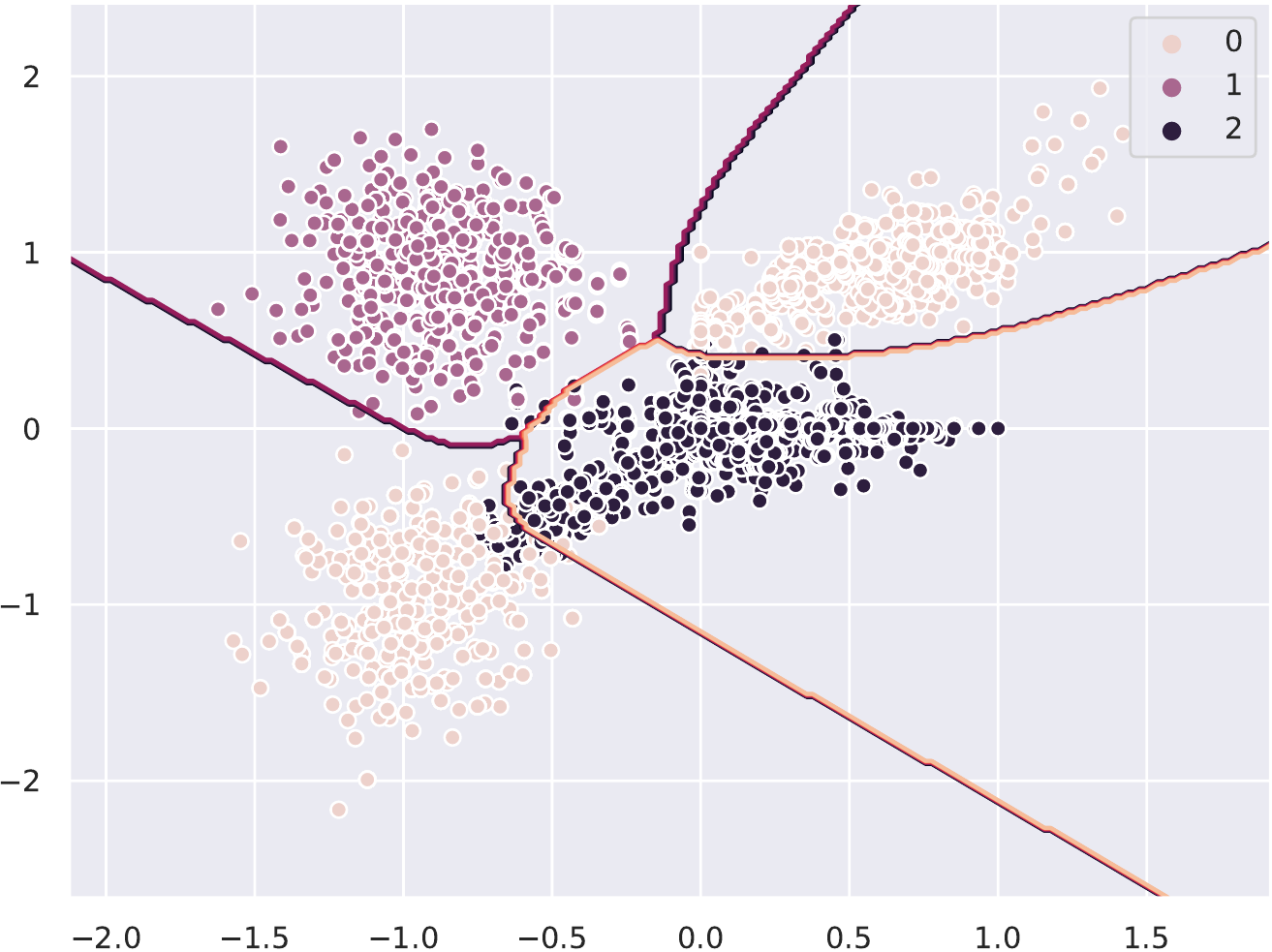}
	\caption{WPGD ($\e = 0.2$)}
	\label{fig:tr_class_w_50}
	\end{subfigure}
	\begin{subfigure}[t]{\sz\textwidth}
	\centering
	\includegraphics[width=\textwidth]{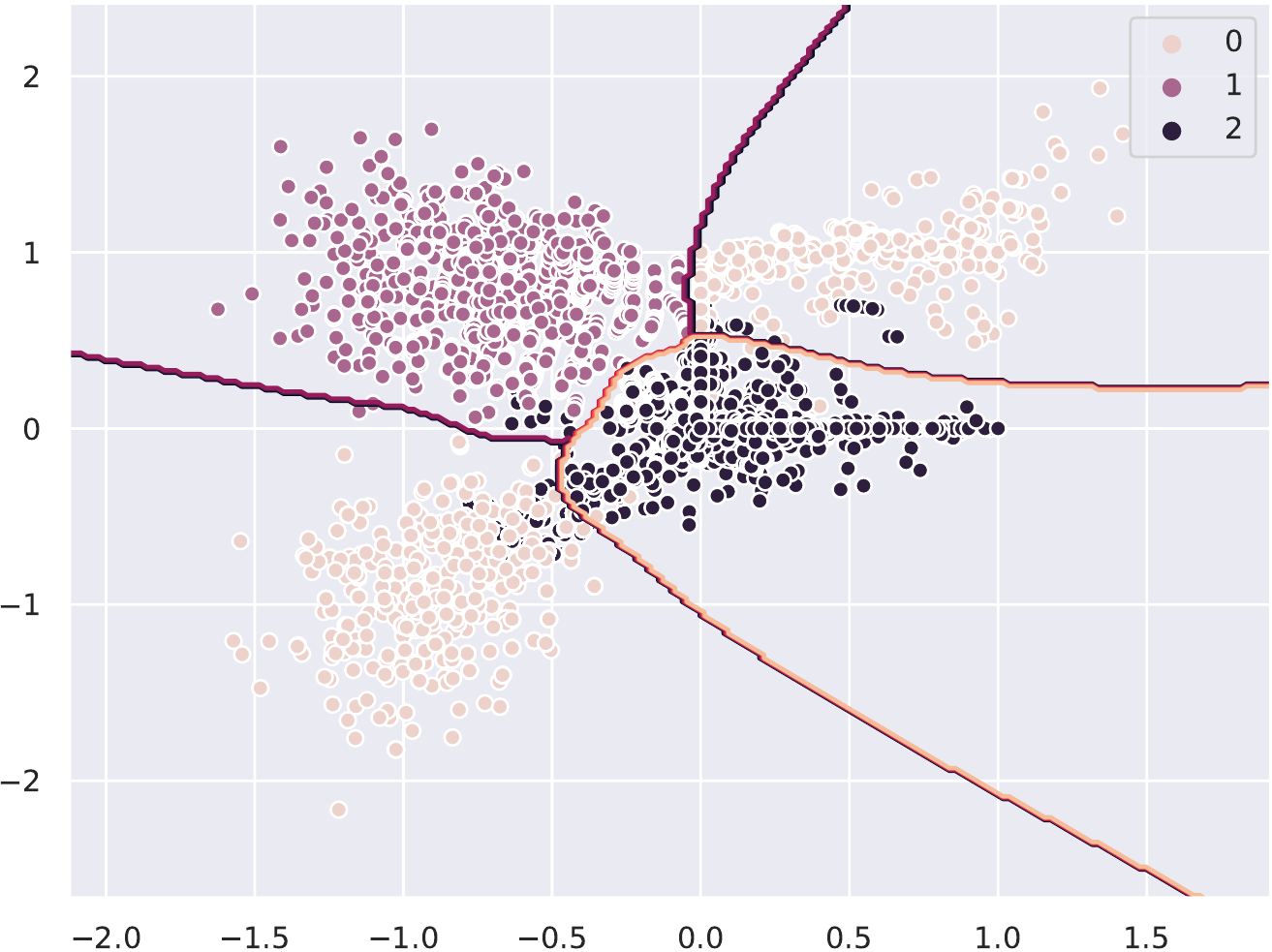}
	\caption{WPGD ($\e = 0.4$)}
	\label{fig:tr_class_w_100}
	\end{subfigure}
	\begin{subfigure}[t]{\sz\textwidth}
	\centering
	\includegraphics[width=\textwidth]{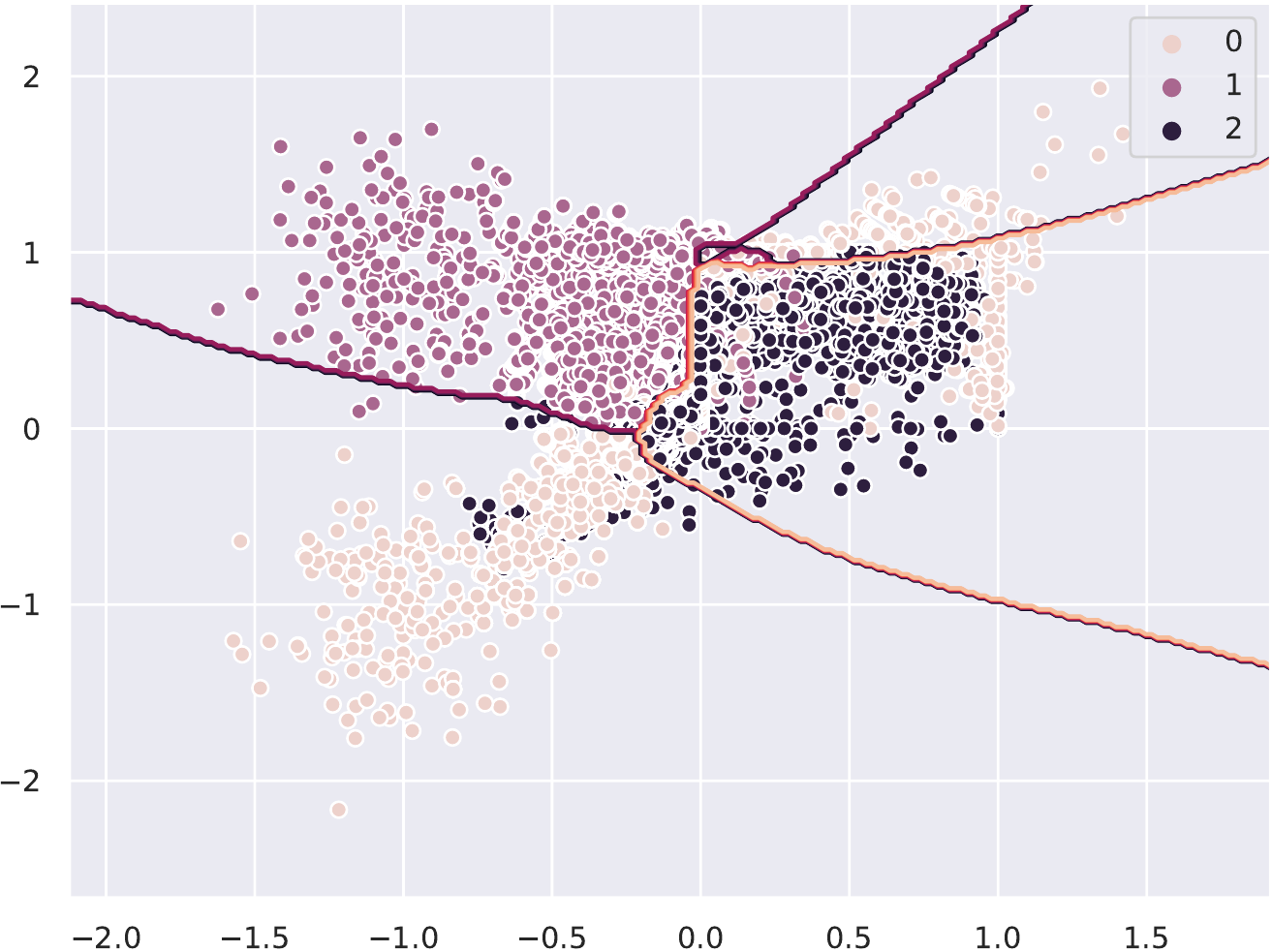}
	\caption{WPGD ($\e = 0.8$)}
	\label{fig:tr_class_w_200}
	\end{subfigure}
\caption{Effect of directional robust $\ell_2$-training on a simple classification problem. 
WPGD flatten the boundaries where the cost is low and preserve them where the cost is high.}
\label{fig:example_WPGD}
\end{figure*}

\begin{remark}
One may find the claim that since visually similar are separated by more complex boundaries, it obviously hurts robustness. However, the range of values of $\e$ used for robust training are much smaller than the minimal distance between two images in the dataset. Thus, at least in principle, it is not still clear why it is not possible to obtain robustness and accuracy at the same time. 
\end{remark}

\section{Wasserstein Projected Gradient Descent}
\label{s:wpgd}
\cref{app:wasserstein} briefly reviews the the necessary background on discrete Optimal transport tools, while~\cref{subsec:wpgd} introduces a new formulation of directional adversarial training.

\subsection{Wasserstein metric and optimal transportation}
\label{app:wasserstein}

The cost between classes, referred to as label metric, is defined in the following:
\begin{definition}[Label metric $\cp$]
\label{def:label_metric}
A symmetric positive semi-definite matrix $C \in \rkk$ defines a pseudo-Riemannian metric on the domain, an entry $C_\kkp$ is the cost of transporting unit probability mass from class $k$ to class $k'$. Note that $C_{k, k} = 0$. The notation $\cp$ denotes the element-wise $p^{\trm{th}}$-power of $C$.
\end{definition}

The other building blocks are the optimal transportation problem~\cite{santambrogio2015optimal} and the Wasserstein metric over probability distributions. 
Given two probability distributions $q, q'$ supported on $K$ classes, the $p$-Wasserstein distance between $q$ and $q'$ for $p \in [1, \infty)$ is defined to be
\beq{
    \wpp \rbrac{q,q'} = \inf_{\T\ \in\ \Pi(q,q')}\ \ag{\T,\ \cp}
    \label{eq:wpp}
}
where $\Pi(q, q') = \cbrac{\T \in \rkk:\  q = \T \ones,\ q' = \T^\top \ones}$ is the set of joint probability distributions with $q$ as the right marginal and $q'$ as the left marginal; $\ones$ denotes the all-one vector and $\ag{\cdot, \cdot}$ is the Frobenius inner product on matrices. The Wasserstein distance is the optimal cost of transporting probability mass from an initial distribution $q$ to a final distribution $q'$. For $0 < p \leq 1$, the Wasserstein distance in~\cref{eq:wpp} is defined to be $\wp \rbrac{q,q'} = \inf_{\T\ \in\ \Pi(q,q')}\ \ag{\T,\ \cp}$; note the absence of $p^{\trm{th}}$ power on the left-hand side. For any separable complete metric space $(\XX, d)$ and $p > 0$, the metric space $(\PP_p, \wp)$ is complete and separable, $\PP_p$ being the set of probability distributions supported on $\XX$~\cite{ambrosio2008gradient}.
%

Problem~\cref{eq:wpp} is called the Kantorovich relaxation~\cite{santambrogio2015optimal} of the original optimal transport problem with $\Pi = \rkk$~\cite{monge1781memoire} and it takes $\OO(K^3)$ operations to solve it using linear programming or interior point methods. \cite{cuturi2013sinkhorn} proposed a smoothed alternative to~\cref{eq:wpp} by adding a convex negative entropic term
\aeq{
    \lwpp \rbrac{q,q'} = \inf_{\T \in \Pi(q,q')}\ \ag{\T, \cp} - \l^{-1} \H(\T), \quad
    \label{eq:lwpp}
}
$\H(\T) = -\sum_{\kkp=1}^K\ \T_\kkp \log \T_\kkp$ that enables an efficient algorithm based on Sinhorn-Knopp iteration~\cite{sinkhorn1964relationship} to approximate $\T^*$. Large values of $\l$ give better approximation to the exact distance $\wpp$ and it can be shown that $\lwpp$ converges to $\wpp$ as $\l \to \infty$~\cite{peyre2018computational}.

Sinhorn-Knopp iteration is a costly algorithm if the number of classes $K$ is large or the metric $\cp$ is complex. However as the following lemma shows, if one of the probability distributions is a one-hot vector, one can compute the optimal transport $\T^*$ in closed form. 
Indeed, in this paper, the $p$-Wasserstein distance is computed between the ground-truth $y(x)$ and the network predictions $\yh(x)$, the former being a one-hot vector.

\begin{lemma}[Closed-form Wasserstein distance]
\label{lem:closed_form_wpp}
For any normalized $q$, if the target probability distribution $q'$ is a one-hot vector, the Wasserstein distance $\wpp$ can be computed in closed form and is given by
\[
    \wpp(q, q') = \cp_{\k^*}\ q
\]
where $\k^* = \argmax_k q'$. The optimal transport is such that its $\rbrac{\k^*}{^\trm{th}}$ column is $q$.
\end{lemma}
The proof of this lemma follows from the observation that the set $\Pi(q, q')$ is degenerate for one-hot $q'$, the constraints $\T^\top \ones = q'$ and $\T \ones = q$ force the $\rbrac{\k^*}{^\trm{th}}$ column of $\T$ to be simply $q$. Note that the Wasserstein distance is symmetric and therefore the same statement holds for $\wpp(q', q)$.
Finally, the regularized~\cite{cuturi2013sinkhorn} Wasserstein Loss is defined as follows:
\begin{definition}[Wasserstein Loss]
\label{def:wasserstein_loss}
The Wasserstein Loss can now be defined as
\beq{
    \lw(\th;\ x) = \cp_{\k(x)}\ \yh(x) - \f{\l^{-1}}{\log K}\ \H(\yh(x));
    \label{eq:ellw}
}
here $\cp_{y(x)}$ denotes the $y(x)^{\trm{th}}$ row of the matrix $\cp \in \rkk$. Note that computing $\lw(\th; x)$ and back-propagating through it has the same computational complexity as standard cross-entropy.
\end{definition}

\subsection{WPGD}
\label{subsec:wpgd}
The saddle point formulation for the Wasserstein loss~\cref{eq:ellw} can be modified to lead to the following definition.
\begin{definition}[Robust Wasserstein loss]
\label{def:rw}
The Robust Wasserstein loss is defined as
\beq{
    \min_\th\ \E_{X}\ \lce(x^*;\ \th), \quad  x^* = \argmax_{ \norm{x'-x}_\infty \leq \e}\ \lw(\th;\ x')
    \label{eq:lws}
}

\end{definition}

The outer loop remains the same while the inner loop is responsible to find the adversarial example which maximize the Wasserstein loss $\lw$. This implies that at the beginning of training WPGD will prefer directions connecting visually distant classes, such as, cat and truck, preventing to flatten regions between similar classes. It is important to note that during training there is an implicit trade-off between choosing directions suggested by the metrics and gradients directions. In fact, the loss gradient is nothing else that an inner product of the $K$ logit's gradients and the the row $k$-th row of $C$.
Imposing an approximation of the real visual metric, helps to efficiently explore the $\ell_\infty$-ball which, especially for high-dimensional input can be hard to explore, leading to better results. For WPGD experiments, the metrics previously described will be used.

\section{Experiments}
This Section provides the experimental findings of the WPGD approach.

\subsection{Datasets and networks}
\label{ss:datasets_networks}
In this paper, the MNIST~\cite{lecun1998gradient}, \Ct, \Ch datasets~\cite{krizhevsky2009learning} and \Ti~\cite{tiny} dataset are used for the experiments. For all datasets, images are normalized to have pixel intensities between $[0,1]$.
 The adversarial vulnerability of neural networks increases with the number of output classes~\cite{fawzi2018adversarial}. In this context, is it worth emphasizing that the \Ti~dataset with $200$ classes is a viable dataset for benchmarking adversarial learning algorithms; this dataset is however less popular in the literature which primarily focuses on MNIST and \Ct.
For the CIFAR datasets, it is used standard data-augmentation which involves mirror flipping with probability of $0.5$ and random crops of size $32\times 32$ after padding images by $4$ pixels on each side. The following networks are used in all the experiments:
\begin{enumerate}[label={(\roman*)}]
\item \tbf{\wrnst:} Wide-Residual network of~\cite{zagoruyko2016wide} with 16 layers, a widening factor of $10$, weight decay of $5 \times 10^{-4}$ and zero dropout.
    \item \tbf{\wrnft:} Wide-Residual network of~\cite{zagoruyko2016wide} with 50 layers, a widening factor of $10$, weight decay of $5 \times 10^{-4}$ and zero dropout.
    \item \tbf{\wrnte:} Wide-Residual network of~\cite{zagoruyko2016wide} with $28$ layers, a widening factor of $10$, weight decay of $5 \times 10^{-4}$ and zero dropout.
\end{enumerate}
All networks are trained with stochastic gradient descent (SGD), Nesterov's momentum of $0.9$ and mini-batch size of $128$. %

\subsection{Algorithms}
The following four algorithms will be compared:
\begin{enumerate}[label={(\roman*)},itemsep=3pt,topsep=0pt,
                parsep=0pt,partopsep=0pt]
    \item $\ce$: This is the standard cross-entropy loss $\lce$ defined in~\cref{eq:ellce}.
    \item  $\rce$: This is the algorithm of~\cite{madry2017towards}; the saddle-point problem~\cref{eq:adversarial_training} is solved with $8$ steps in the inner loop to compute the adversarial image.
    \item $\rw$: This is the robust Wasserstein loss described in~\cref{def:rw} where the inner loop in PGD searches over the adversarial image that maximizes the Wasserstein transport cost. The computational complexity of WPGD is the same as that of PGD. WPGD is compared with three different value of $p=1,2.5,10$.
\end{enumerate}

W-$s$-10 represents the wideresnet architecture with $s$ layers.
In order to test robustness, 20-steps PGD attacks are performed starting from a random (uniformly sampled) position inside the $\ell_\infty$ ball of the test image $x$.
All the WPGD experiments are run with the cost matrix provided by the WordNet metric~\cite{miller1995wordnet}.
\subsection{Directional robustness of WPGD}

\begin{figure}[htpb]
    \centering
    \includegraphics[width=0.49\textwidth]{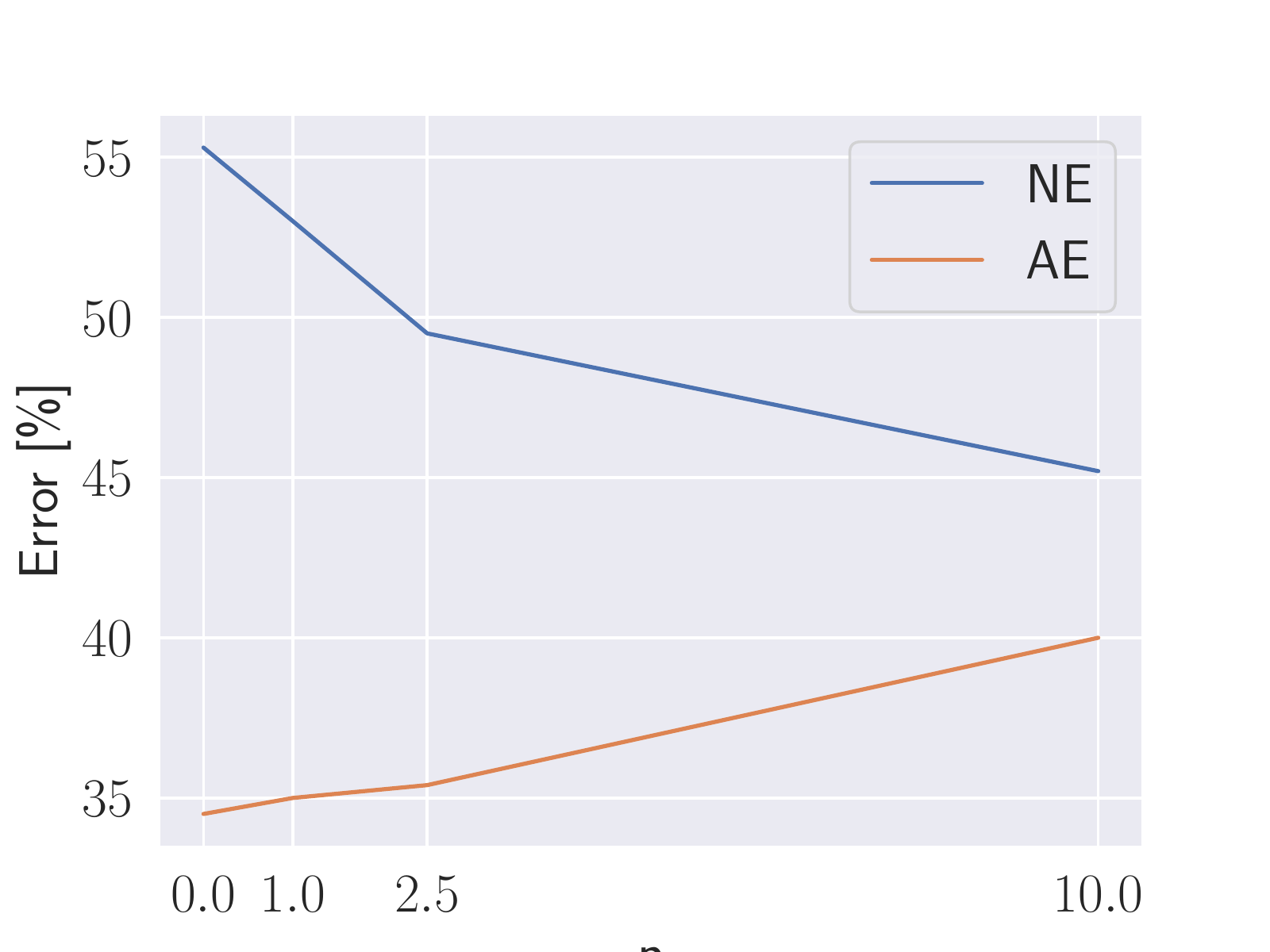}
     \includegraphics[width=0.49\textwidth]{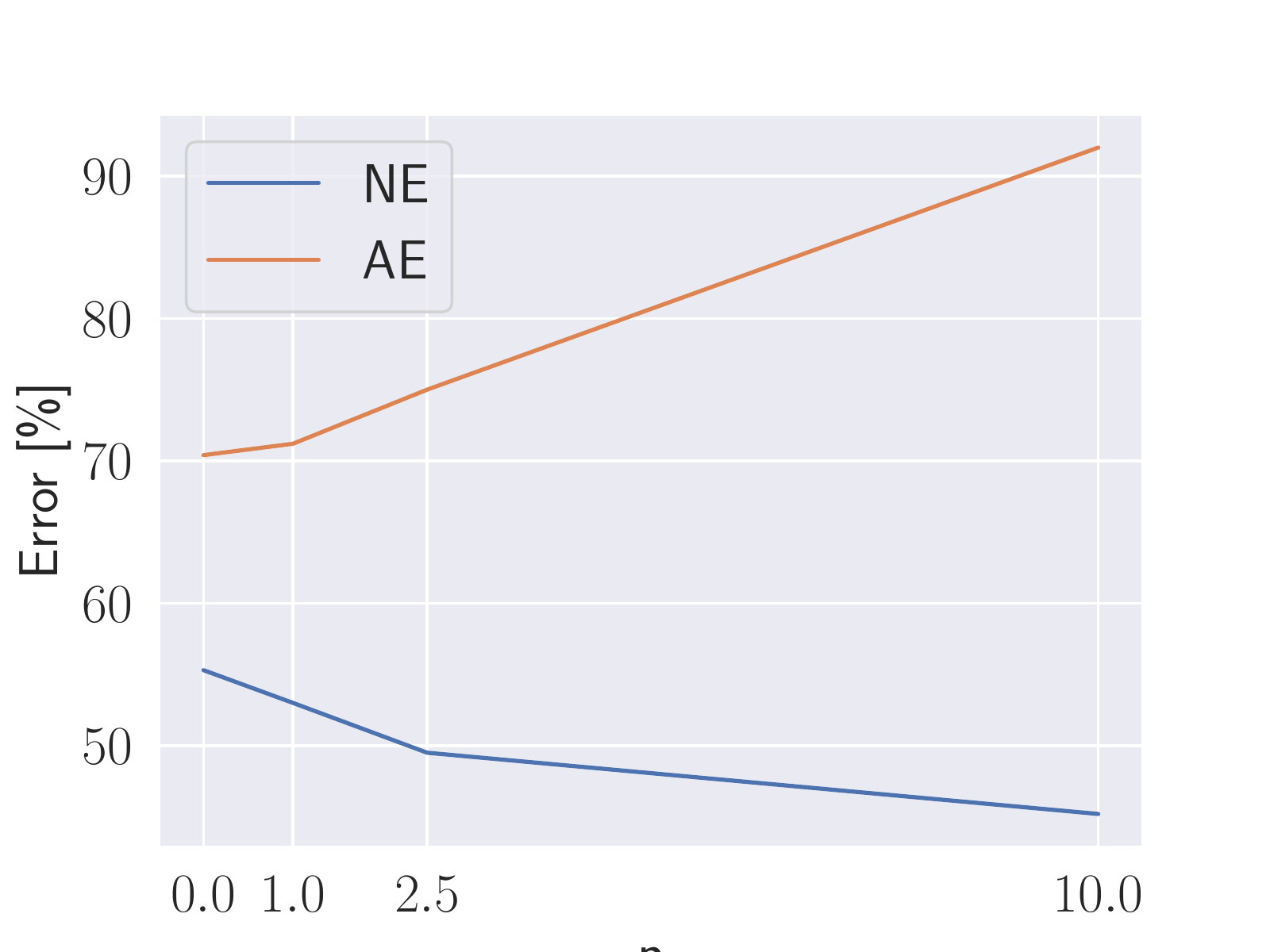}
    \caption{Accuracy \& robustness trade-off: Results for \wrnst~and $\e=4$, for \Ct~(left) and \Ti~(right). Increasing $p$ (x-axis), enables to improve accuracy on fine-grained classification at the price of robustness on pairs of similar classes}
    \label{fig:tradeoff}
\end{figure}

\begin{figure}[htbp]
	\centering
	\captionsetup[subfigure]{justification=centering}
	\begin{subfigure}[b]{0.48\textwidth}
	        \centering
	        \includegraphics[width=\textwidth]{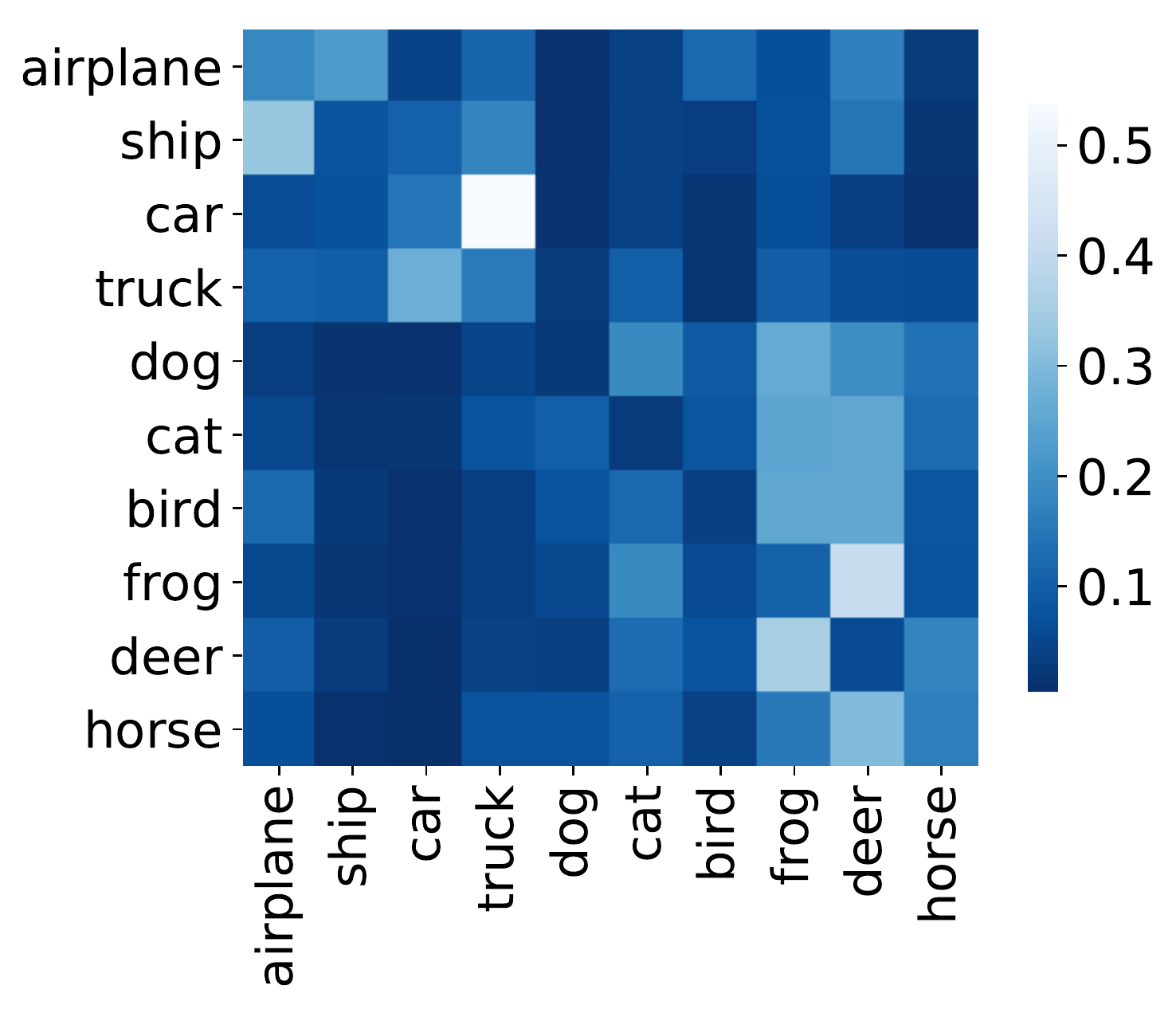}
	        \caption{$\rce$}
	        \label{fig:attack_wrn_c10_pgd_mat}
	      \end{subfigure}
  \begin{subfigure}[b]{0.48\textwidth}
	        \centering
	        \includegraphics[width=\textwidth]{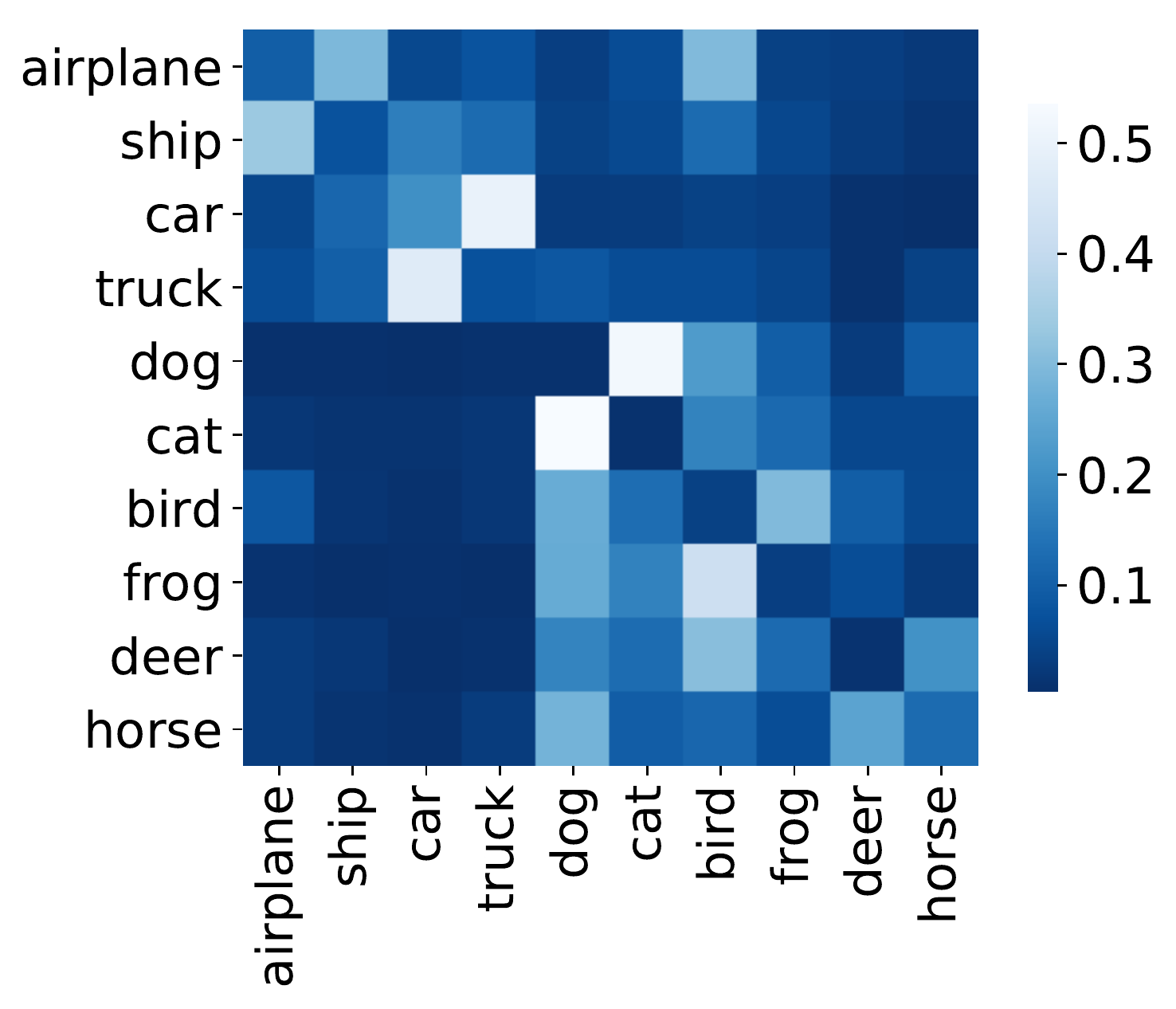}
	        \caption{$\rw$}
	        \label{fig:attack_wrn_c10_wpgd_mat}
	    \end{subfigure}
	    \caption{\Ct. Characterization of adversarial robustness for WPGD and PGD defenses. Applied perturbations have norm $\e = 16$ on~\wrnte. 
	    WPGD is performed using $p=2.5$.~\cref{fig:attack_wrn_c10_wpgd_mat} shows the WPGD obtains directionally robustness: in this case the network is more robust for perturbations between two visually different classes.}
	    \label{fig:rob_pgd_vs_wpgd_C10}
\end{figure}

In~\cref{tab:summary} the main results of natural training (CE) and robust training (PGD) for \Ct~and \Ti are reported. ~\cref{app:wpgd_res} reports a summary table for quantitative results on directional robustness. 
Instead, in~\cref{fig:tradeoff} it is shown the trade-off arising from WPGD training. As $p$ increases, fine-grained classification is more preserved. In addition to standard accuracy, the characterizations of adversarial robustness of PGD and WPGD is compared. In~\cref{fig:rob_pgd_vs_wpgd_C10} it shown that WPGD-trained networks with a strong metric tend to be more robust between visually distant classes, which supports out claims. 
For sake of clarity, only results for \Ct~and \Ti and \wrnst~are reported. 
Interestingly, WPGD is less robust than PGD for classes \emph{bird} and \emph{airplane}: thus, imposing a metric, even if it is only approximately correct, seems to help to obtain more visually meaningful errors.

\begin{table*}[!htpb]
\centering
\renewcommand{\arraystretch}{1}
\begin{tabularx}{0.49\textwidth}{X X X X X X X}
\toprule
\tbf{C10} & \multicolumn{2}{c}{$\ce$}  & \multicolumn{2}{c}{$\rce$}\\
& \rsz{16-10} & \rsz{28-10} & \rsz{16-10} & \rsz{28-10}\\
 \midrule
NE & 4.4 & 3.9 & 14.11 & 13.9\\
AE & 100 & 100 & 34.5& 31.25\\
\bottomrule
\end{tabularx}
\begin{tabularx}{0.49\textwidth}{X X X X X X X}
\toprule
\tbf{Tiny} & \multicolumn{2}{c}{$\ce$}  & \multicolumn{2}{c}{$\rce$}\\
& \rsz{16-10} & \rsz{28-10} & \rsz{16-10} & \rsz{28-10}\\
 \midrule
NE & 37.7 & 36.9 & 55.3 & 36.9 \\
AE  & 99.9 & 100 & 70.4 & 70.5\\
\bottomrule
\end{tabularx}

\caption{Summary of errors in [\%] for \wrnst~and \wrnte, with $\e = 4$ and $k=20$ under $\ell_\infty$ perturbations.}
\label{tab:summary}
\end{table*}

\newpage
\subsection{Supplementary comparisons for CE, PGD and WPGD}
\label{app:wpgd_res}
\begin{figure*}[htbp]
    \centering
    \includegraphics[width=0.32\textwidth]{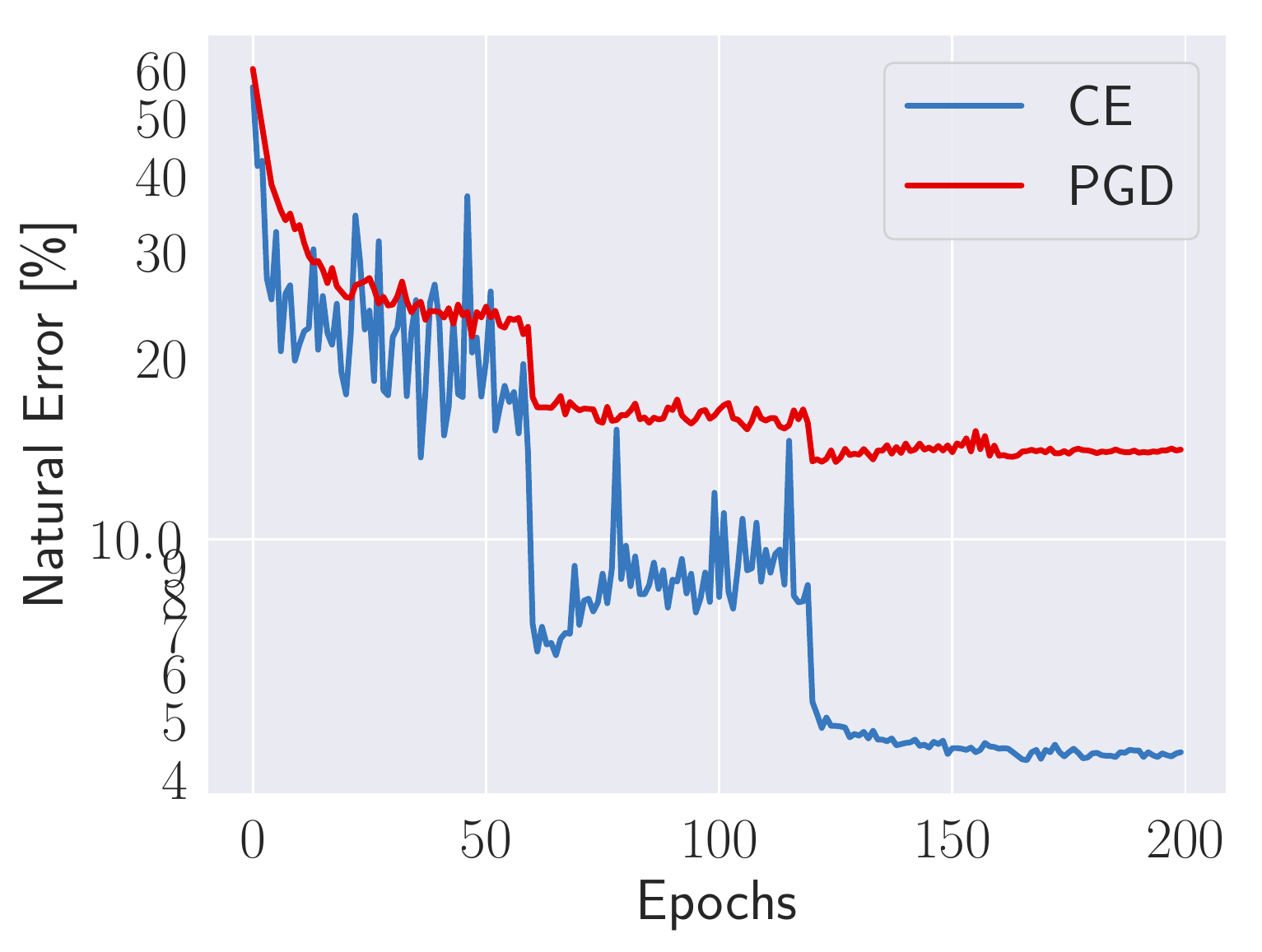}
    \includegraphics[width=0.32\textwidth]{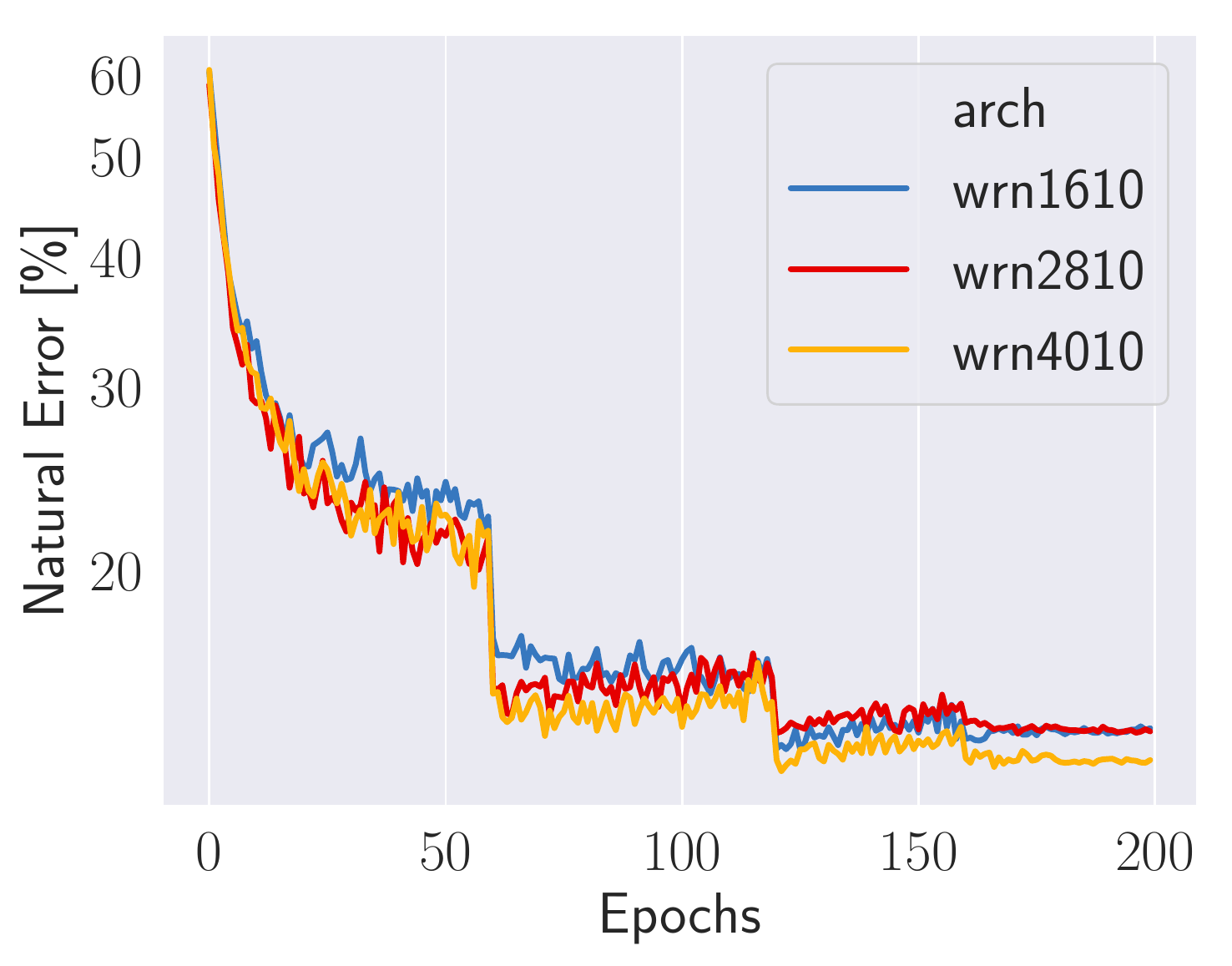}
    \includegraphics[width=0.32\textwidth]{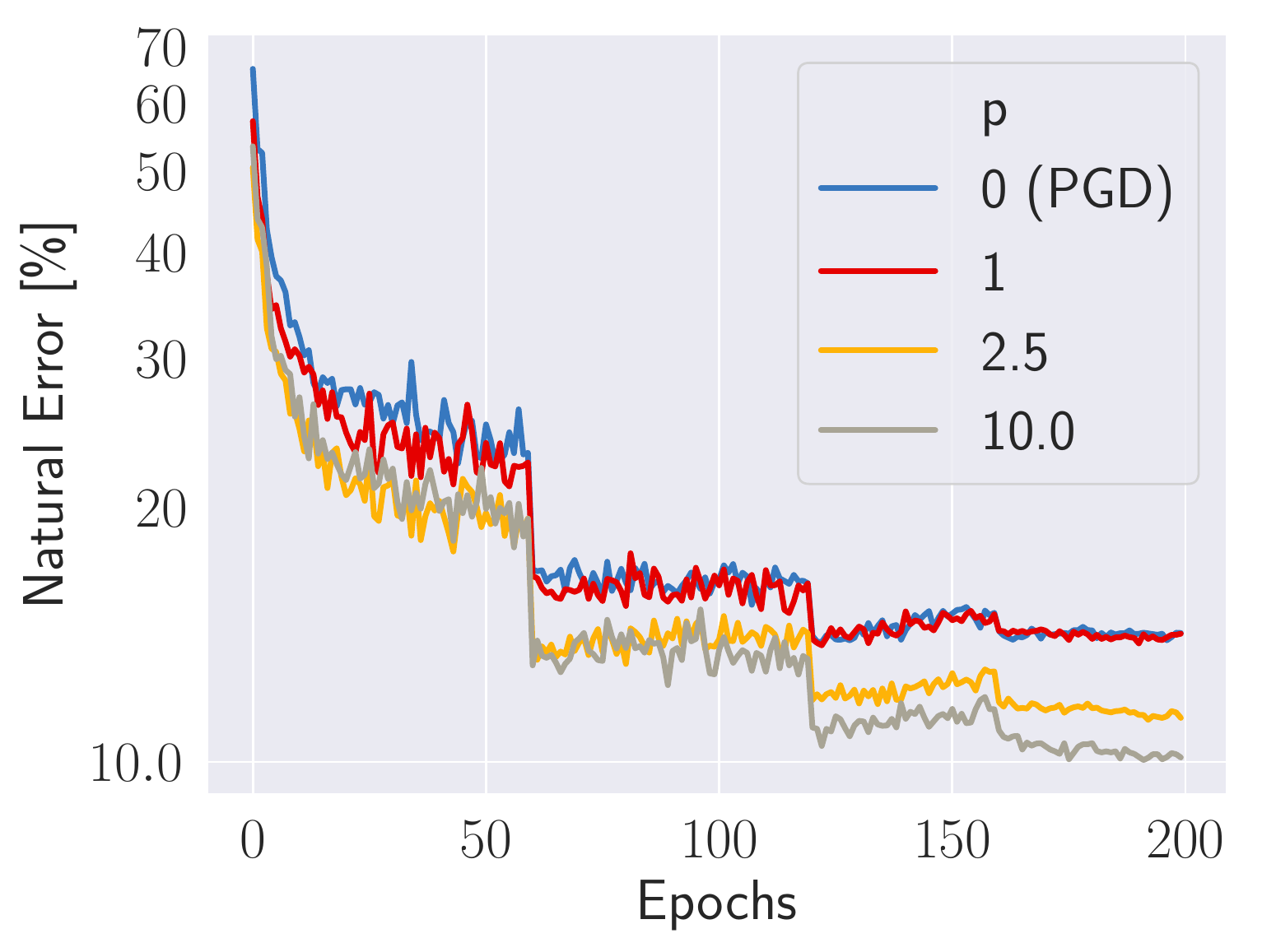}
    \caption{\Ct. Comparisons: (Left) standard training vs PGD; (Middle) different architectures on robust training; (Right) PGD vs WPGD. WPGD is slightly better in terms of accuracy.}
   \label{fig:cifar10_comparisons}
\end{figure*}

~\cref{fig:cifar10_comparisons} report curves plot for PGD and WPGD for \Ct and \Ti. Moreover,~\cref{tab:summary_robustness} reports the summary of weighted robustness score $S$ defined as:
\beq{
\label{eq:score}
S = \sum_{i,j} c_{i,j} m_{i,j}
}
where $M = \{m_{ij}\}_{i,j=1}^K$ is the adversarial confusion matrix, $C = \{c_{ij}\}_{i,j=1}^K$ is the metric of the given dataset.
Attacks are computed maximizing the loss~\cref{eq:lws}, that is considering the worst-case scenario in which the attacker knows the metric.
This score weighs more errors in correspondence of high cost.
In order to make results legible, the zero reference is set to the PGD-trained model. As it can be seen increasing $p$, results in reducing the score $S$, which means that, on average, more similar classes are reached.

\begin{figure}[htbp]
    \centering
    \includegraphics[width=0.32\textwidth]{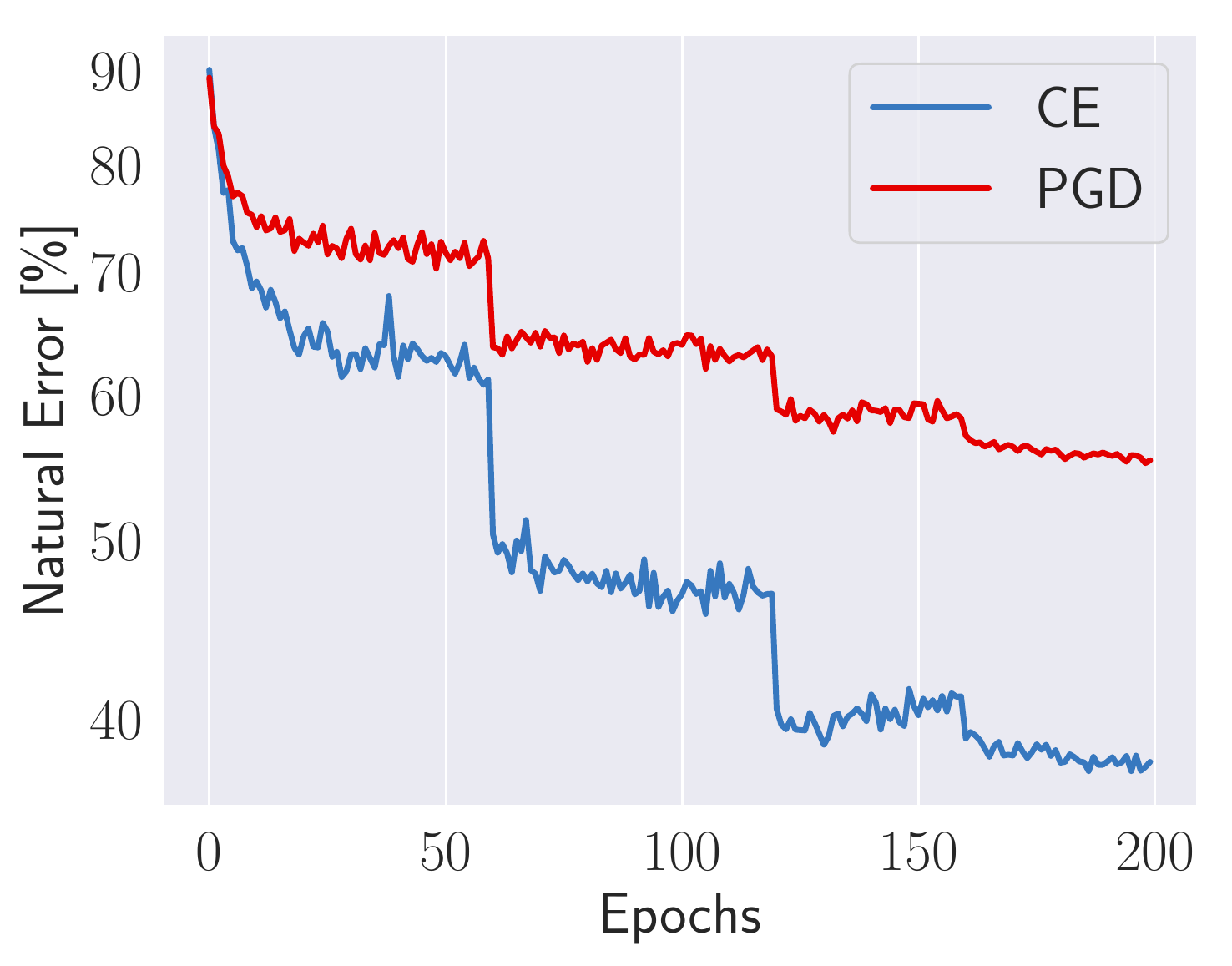}
     \includegraphics[width=0.32\textwidth]{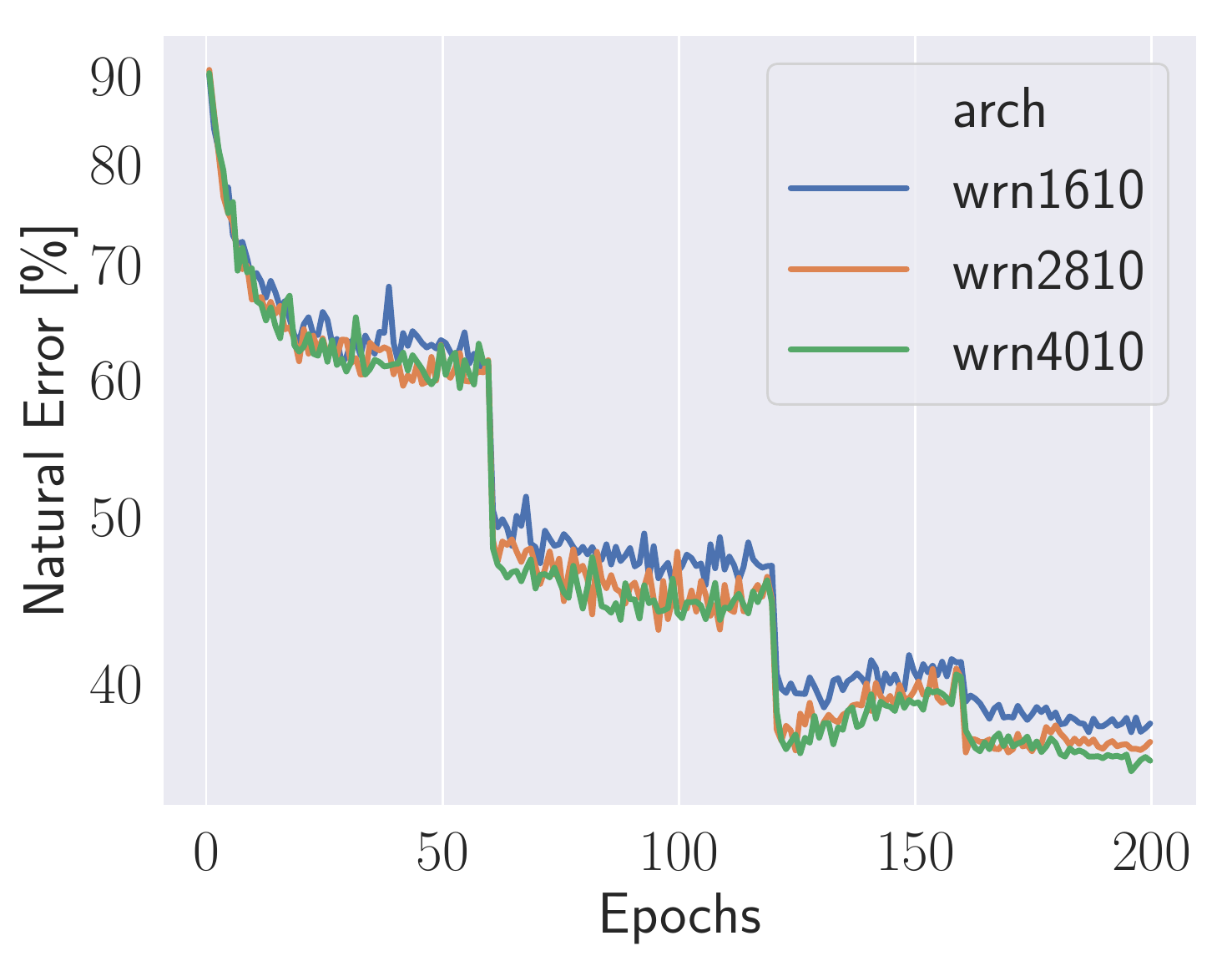}
    \includegraphics[width=0.32\textwidth]{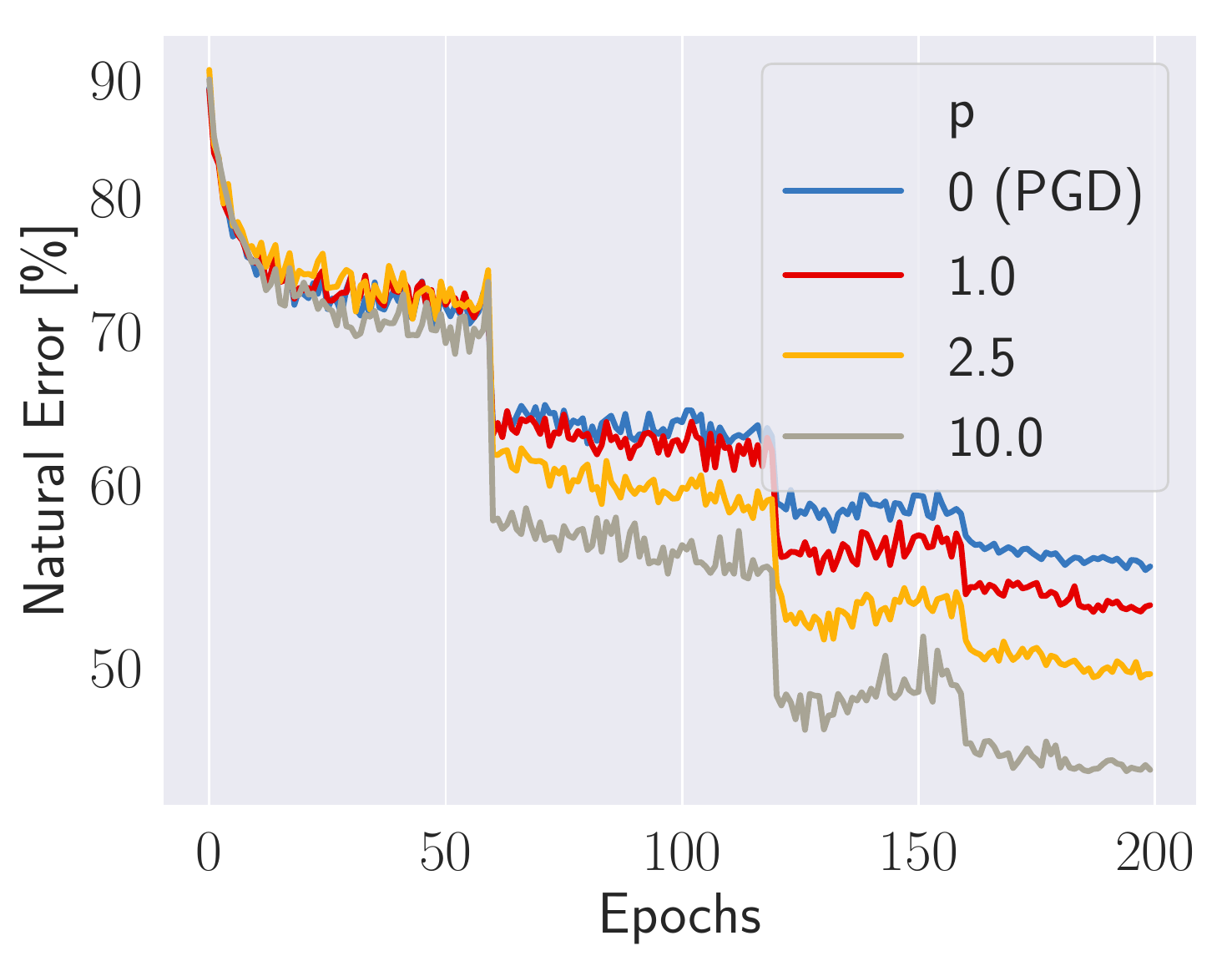}
    \caption{\Ti. Comparisons: (Top) standard training vs PGD; (Middle) different architectures on robust training; (Bottom) PGD vs WPGD. WPGD is slightly better in terms of accuracy.}
    \label{fig:tiny_comparisons}
\end{figure}

\begin{table}[htbp]
    \centering
\begin{tabular}{cccccc}
\toprule
     &  \tbf{AE [$\%$]} &  \tbf{$p$} &  \tbf{dataset} &   \tbf{$S$} \\
\midrule    
\wrnst &  34.53 &    0.0 &         \Ct &      -0.14 \\
\wrnst &  34.62 &    1.0 &         \Ct &      -0.26\\
\wrnst &  34.98 &    2.5 &         \Ct &     -0.34 \\
\wrnst &  39.76 &   10.0 &         \Ct &      -0.53 \\
 \wrnte &  31.24 &    0.0 &         \Ct &      0.00 \\
\wrnst &  70.23 &    1.0 &  \Ti &      -6.33 \\
\wrnst &  73.61 &    2.5 &  \Ti &    -12.45 \\
\wrnst &  92.62 &   10.0 &  \Ti &     -55.17 \\
\wrnte &  69.84 &    1.0 &  \Ti &     -9.62 \\
\wrnte &  69.69 &    0.0 &  \Ti &    -9.48 \\
\bottomrule
\end{tabular}
    \caption{Summary of weighted robustness score $S$ defined in~\cref{eq:score} for $\e=4$. In order to make results more legible, the zero reference (for each dataset) is set to the PGD-trained model. As it can be seen, increasing $p$ results in reducing the score $S$, which means that, on average, more similar classes are reached.}
    \label{tab:summary_robustness}
\end{table}

\newpage
\section{Related work}
\label{s:related_work}
This work is related to~\cite{madry2017towards,Tsipras:2018aa}. Although they give theoretical and practical results on the connection between robustness and accuracy for adversarial training, they don't analyze how the accuracy gap is distributed. They also argue that adversarial training requires extra capacity in order to build complex boundaries~\cite{kolter2017provable}. In contrast,~\cite{Moosavi-Dezfooli:2018aa} have recently argued that adversarial training leads to flatter decision boundaries and in fact, explicitly penalizing the curvature of the decision boundary is a good technique to train robust classifiers. 
Results in this paper corroborate these findings. 
The accuracy gap of adversarially trained networks with respect to standard cross-entropy trained networks can be explained, very well  the experiments show, by the network getting these pairs of classes incorrect.
Semantic metrics, e.g., those derived from WordNet~\cite{miller1995wordnet} to aid visual classification have been popular to introduce a new data-modality in standard supervised learning~\cite{Deng_2010, deng2014large}. 
This paper identifies the inherent visual metric that the network induces while being trained using cross-entropy loss or the adversarial loss.
Lastly, using an optimal transport formulation to impose a metric on the label space of deep networks bears close resemblance to the work of~\cite{frogner2015learning}. 
This work uses the Wasserstein loss computed using the Sinhorn-Knopp iteration to predict multi-label images. The present paper is the first to use the optimal transport formulation to induce a cost-sensitive adversarial training of deep networks. Further, for single-label images, it shown that the optimal transport problem has a closed form solution which makes it computationally equivalent to the cross-entropy loss; this simple but powerful property may be of independent interest for problems like hierarchical classification~\cite{girshick2014rich,wu2017hierarchical,bagherinezhad2018label}.

\section{Conclusions and future work}
\label{s:discussion}

While the literature on adversarial training is flourishing, profound studies towards understanding its implication and sensitivity to common real-world applications are still lacking. In particular, this paper focused on applications that are cost-sensitive or the dataset is unbalanced. Moreover, due to an intrinsic trade-off between robustness and accuracy, it is of paramount importance to be to govern such trade-off when designing and implementing machine and deep learning-based applications where a certain amount of accuracy is required.
In liue of this, the present paper made several advances towards understanding better robustness from one side and being able to semantically control it from the other side. 

In particular, this paper identified that the accuracy gap in adversarial training comes from the loss of fine-grained classification capabilities in neural networks. This observation motivates the optimal transport formulation: a metric on the label space that measures the distance to the boundary for standard cross-entropy training or, often equivalently, a semantic metric obtained from external data modalities such as WordNet, reduces the search space and makes it easier to discover---and fix---these classes during adversarial training, resulting in an improvement of accuracy at the cost of (directional) robustness. 
It is conceivable that, although a high-dimensional classifier may always remain vulnerable to adversarial perturbations, it is possible to build robust, real-world systems by incorporating such diverse data. Thus, this work is a first step toward a principled robust training for real-world applications involving artificial intelligence and deep learning.

Future works will regard the study of methodologies or heuristics to systematically control the robustness-accuracy trade-off without the need of tuning $\e$ by hyper-parameter tuning. Moreover, another future direction of research is the application of the WPGD approach to other problems like fraud detection and Predictive Maintenance.

\section*{Acknowledgments}
The authors gratefully acknowledge the support of NVIDIA Corporation with the donation of the Titan V GPU used for this research and Amazon Web Services for donating research credits.


%


\begin{thebibliography}{10}

\bibitem{tiny}
Tinyimagenet.
\newblock \url{https://tiny-imagenet.herokuapp.com/}.

\bibitem{ambrosio2008gradient}
Luigi Ambrosio, Nicola Gigli, and Giuseppe Savar{\'e}.
\newblock {\em Gradient flows: in metric spaces and in the space of probability
  measures}.
\newblock Springer Science \& Business Media, 2008.

\bibitem{athalye2018obfuscated}
Anish Athalye, Nicholas Carlini, and David Wagner.
\newblock Obfuscated gradients give a false sense of security: Circumventing
  defenses to adversarial examples.
\newblock {\em arXiv:1802.00420}, 2018.

\bibitem{bagherinezhad2018label}
Hessam Bagherinezhad, Maxwell Horton, Mohammad Rastegari, and Ali Farhadi.
\newblock Label refinery: Improving imagenet classification through label
  progression.
\newblock {\em arXiv:1805.02641}, 2018.

\bibitem{beke2019learning}
Aykut Beke and Tufan Kumbasar.
\newblock Learning with type-2 fuzzy activation functions to improve the
  performance of deep neural networks.
\newblock {\em Engineering Applications of Artificial Intelligence},
  85:372--384, 2019.

\bibitem{boyd2004convex}
Stephen Boyd and Lieven Vandenberghe.
\newblock {\em Convex optimization}.
\newblock Cambridge university press, 2004.

\bibitem{carlini2017adversarial}
Nicholas Carlini and David Wagner.
\newblock Adversarial examples are not easily detected: Bypassing ten detection
  methods.
\newblock In {\em Proceedings of the 10th ACM Workshop on Artificial
  Intelligence and Security}, pages 3--14. ACM, 2017.

\bibitem{cuturi2013sinkhorn}
Marco Cuturi.
\newblock Sinkhorn distances: Lightspeed computation of optimal transport.
\newblock In {\em Advances in neural information processing systems}, pages
  2292--2300, 2013.

\bibitem{Deng_2010}
Jia Deng, Alexander~C. Berg, Kai Li, and Li~Fei-Fei.
\newblock What does classifying more than 10,000 image categories tell us?
\newblock {\em Lecture Notes in Computer Science}, pages 71--84, 2010.

\bibitem{deng2014large}
Jia Deng, Nan Ding, Yangqing Jia, Andrea Frome, Kevin Murphy, Samy Bengio, Yuan
  Li, Hartmut Neven, and Hartwig Adam.
\newblock Large-scale object classification using label relation graphs.
\newblock {\em Lecture Notes in Computer Science}, pages 48--64, 2014.

\bibitem{deng2009imagenet}
Jia Deng, Wei Dong, Richard Socher, Li-Jia Li, Kai Li, and Li~Fei-Fei.
\newblock Imagenet: A large-scale hierarchical image database.
\newblock In {\em Computer Vision and Pattern Recognition, 2009. CVPR 2009.
  IEEE Conference on}, pages 248--255. Ieee, 2009.

\bibitem{esteva2019guide}
Andre Esteva, Alexandre Robicquet, Bharath Ramsundar, Volodymyr Kuleshov, Mark
  DePristo, Katherine Chou, Claire Cui, Greg Corrado, Sebastian Thrun, and Jeff
  Dean.
\newblock A guide to deep learning in healthcare.
\newblock {\em Nature medicine}, 25(1):24, 2019.

\bibitem{fawzi2018adversarial}
Alhussein Fawzi, Hamza Fawzi, and Omar Fawzi.
\newblock Adversarial vulnerability for any classifier.
\newblock {\em arXiv:1802.08686}, 2018.

\bibitem{fawzi2018analysis}
Alhussein Fawzi, Omar Fawzi, and Pascal Frossard.
\newblock Analysis of classifiers robustness to adversarial perturbations.
\newblock {\em Machine Learning}, 107(3):481--508, 2018.

\bibitem{frogner2015learning}
Charlie Frogner, Chiyuan Zhang, Hossein Mobahi, Mauricio Araya, and Tomaso~A
  Poggio.
\newblock Learning with a wasserstein loss.
\newblock In {\em Advances in Neural Information Processing Systems}, pages
  2053--2061, 2015.

\bibitem{gao2018ima}
Zehai Gao, Cunbao Ma, Yige Luo, and Zhiyue Liu.
\newblock Ima health state evaluation using deep feature learning with quantum
  neural network.
\newblock {\em Engineering Applications of Artificial Intelligence},
  76:119--129, 2018.

\bibitem{girshick2014rich}
Ross Girshick, Jeff Donahue, Trevor Darrell, and Jitendra Malik.
\newblock Rich feature hierarchies for accurate object detection and semantic
  segmentation.
\newblock 11 2013.

\bibitem{goodfellow2014explaining}
Ian~J Goodfellow, Jonathon Shlens, and Christian Szegedy.
\newblock Explaining and harnessing adversarial examples.
\newblock {\em arXiv:1412.6572}, 2014.

\bibitem{guo2017calibration}
Chuan Guo, Geoff Pleiss, Yu~Sun, and Kilian~Q Weinberger.
\newblock On calibration of modern neural networks.
\newblock In {\em Proceedings of the 34th International Conference on Machine
  Learning-Volume 70}, pages 1321--1330. JMLR. org, 2017.

\bibitem{ibitoye2019analyzing}
Olakunle Ibitoye, Omair Shafiq, and Ashraf Matrawy.
\newblock Analyzing adversarial attacks against deep learning for intrusion
  detection in iot networks.
\newblock {\em arXiv preprint arXiv:1905.05137}, 2019.

\bibitem{kannan2018adversarial}
Harini Kannan, Alexey Kurakin, and Ian Goodfellow.
\newblock Adversarial logit pairing.
\newblock {\em arXiv preprint arXiv:1803.06373}, 2018.

\bibitem{kolter2017provable}
J~Zico Kolter and Eric Wong.
\newblock Provable defenses against adversarial examples via the convex outer
  adversarial polytope.
\newblock {\em arXiv preprint arXiv:1711.00851}, 2017.

\bibitem{krizhevsky2009learning}
A.~Krizhevsky.
\newblock Learning multiple layers of features from tiny images.
\newblock Master's thesis, Computer Science, University of Toronto, 2009.

\bibitem{krizhevsky2012imagenet}
Alex Krizhevsky, Ilya Sutskever, and Geoffrey~E Hinton.
\newblock Imagenet classification with deep convolutional neural networks.
\newblock In {\em Advances in neural information processing systems}, pages
  1097--1105, 2012.

\bibitem{lecun1998gradient}
Y.~LeCun, L.~Bottou, Y.~Bengio, and P.~Haffner.
\newblock Gradient-based learning applied to document recognition.
\newblock {\em Proceedings of the IEEE}, 86(11):2278--2324, 1998.

\bibitem{lecun2015deep}
Yann LeCun, Yoshua Bengio, and Geoffrey Hinton.
\newblock Deep learning.
\newblock {\em Nature}, 521(7553):436--444, 2015.

\bibitem{maaten2008visualizing}
Laurens van~der Maaten and Geoffrey Hinton.
\newblock Visualizing data using t-sne.
\newblock {\em Journal of machine learning research}, 9(Nov):2579--2605, 2008.

\bibitem{madry2017towards}
Aleksander Madry, Aleksandar Makelov, Ludwig Schmidt, Dimitris Tsipras, and
  Adrian Vladu.
\newblock Towards deep learning models resistant to adversarial attacks.
\newblock {\em arXiv:1706.06083}, 2017.

\bibitem{maggipinto2018computer}
Marco Maggipinto, Matteo Terzi, Chiara Masiero, Alessandro Beghi, and
  Gian~Antonio Susto.
\newblock A computer vision-inspired deep learning architecture for virtual
  metrology modeling with 2-dimensional data.
\newblock {\em IEEE Transactions on Semiconductor Manufacturing},
  31(3):376--384, 2018.

\bibitem{miller1995wordnet}
George~A Miller.
\newblock {WordNet: a lexical database for English}.
\newblock {\em Communications of the ACM}, 38(11):39--41, 1995.

\bibitem{monge1781memoire}
Gaspard Monge.
\newblock M{\'e}moire sur la th{\'e}orie des d{\'e}blais et des remblais.
\newblock {\em Histoire de l'Acad{\'e}mie Royale des Sciences de Paris}, 1781.

\bibitem{moosavi2017universal}
Seyed-Mohsen Moosavi-Dezfooli, Alhussein Fawzi, Omar Fawzi, and Pascal
  Frossard.
\newblock Universal adversarial perturbations.
\newblock {\em arXiv:1610.08401}, 2017.

\bibitem{Moosavi-Dezfooli:2018aa}
Seyed-Mohsen Moosavi-Dezfooli, Alhussein Fawzi, Jonathan Uesato, and Pascal
  Frossard.
\newblock Robustness via curvature regularization, and vice versa.
\newblock 11 2018.

\bibitem{nassif2019speech}
Ali~Bou Nassif, Ismail Shahin, Imtinan Attili, Mohammad Azzeh, and Khaled
  Shaalan.
\newblock Speech recognition using deep neural networks: A systematic review.
\newblock {\em IEEE Access}, 7:19143--19165, 2019.

\bibitem{papernot2016transferability}
Nicolas Papernot, Patrick McDaniel, and Ian Goodfellow.
\newblock Transferability in machine learning: from phenomena to black-box
  attacks using adversarial samples.
\newblock {\em arXiv:1605.07277}, 2016.

\bibitem{papernot2016practical}
Nicolas Papernot, Patrick McDaniel, Ian Goodfellow, Somesh Jha, Z~Berkay Celik,
  and Ananthram Swami.
\newblock Practical black-box attacks against deep learning systems using
  adversarial examples.
\newblock {\em arXiv:1602.02697}, 2016.

\bibitem{peyre2018computational}
Gabriel Peyr{\'e} and Marco Cuturi.
\newblock Computational optimal transport.
\newblock {\em arXiv:1803.00567}, 2018.

\bibitem{qayyum2019securing}
Adnan Qayyum, Muhammad Usama, Junaid Qadir, and Ala Al-Fuqaha.
\newblock Securing connected \& autonomous vehicles: Challenges posed by
  adversarial machine learning and the way forward.
\newblock {\em arXiv preprint arXiv:1905.12762}, 2019.

\bibitem{qi2018rotor}
Xing Qi.
\newblock Rotor resistance and excitation inductance estimation of an induction
  motor using deep-q-learning algorithm.
\newblock {\em Engineering Applications of Artificial Intelligence}, 72:67--79,
  2018.

\bibitem{rafique2018nonconvex}
Hassan Rafique, Mingrui Liu, Qihang Lin, and Tianbao Yang.
\newblock Non-convex min-max optimization: Provable algorithms and applications
  in machine learning, 2018.

\bibitem{santambrogio2015optimal}
Filippo Santambrogio.
\newblock Optimal transport for applied mathematicians.
\newblock {\em Birk{\"a}user, NY}, 2015.

\bibitem{saxe2019mathematical}
Andrew~M Saxe, James~L McClelland, and Surya Ganguli.
\newblock A mathematical theory of semantic development in deep neural
  networks.
\newblock {\em Proceedings of the National Academy of Sciences},
  116(23):11537--11546, 2019.

\bibitem{schmidt2018adversarially}
Ludwig Schmidt, Shibani Santurkar, Dimitris Tsipras, Kunal Talwar, and
  Aleksander Madry.
\newblock Adversarially robust generalization requires more data.
\newblock {\em arXiv:1804.11285}, 2018.

\bibitem{sinha2018certifying}
Aman Sinha, Hongseok Namkoong, and John Duchi.
\newblock Certifying some distributional robustness with principled adversarial
  training.
\newblock 2018.

\bibitem{sinkhorn1964relationship}
Richard Sinkhorn.
\newblock A relationship between arbitrary positive matrices and doubly
  stochastic matrices.
\newblock {\em The annals of mathematical statistics}, 35(2):876--879, 1964.

\bibitem{susto2014machine}
Gian~Antonio Susto, Andrea Schirru, Simone Pampuri, Se{\'a}n McLoone, and
  Alessandro Beghi.
\newblock Machine learning for predictive maintenance: A multiple classifier
  approach.
\newblock {\em IEEE Transactions on Industrial Informatics}, 11(3):812--820,
  2014.

\bibitem{szegedy2013intriguing}
Christian Szegedy, Wojciech Zaremba, Ilya Sutskever, Joan Bruna, Dumitru Erhan,
  Ian Goodfellow, and Rob Fergus.
\newblock Intriguing properties of neural networks.
\newblock {\em arXiv:1312.6199}, 2013.

\bibitem{Tsipras:2018aa}
Dimitris Tsipras, Shibani Santurkar, Logan Engstrom, Alexander Turner, and
  Aleksander Madry.
\newblock Robustness may be at odds with accuracy.
\newblock 05 2018.

\bibitem{tsipras2018there}
Dimitris Tsipras, Shibani Santurkar, Logan Engstrom, Alexander Turner, and
  Aleksander Madry.
\newblock There is no free lunch in adversarial robustness (but there are
  unexpected benefits).
\newblock {\em arXiv:1805.12152}, 2018.

\bibitem{tsipras2018robustness}
Dimitris Tsipras, Shibani Santurkar, Logan Engstrom, Alexander Turner, and
  Aleksander Madry.
\newblock Robustness may be at odds with accuracy.
\newblock In {\em International Conference on Learning Representations}, 2019.

\bibitem{wu2017hierarchical}
Cinna Wu, Mark Tygert, and Yann LeCun.
\newblock Hierarchical loss for classification.
\newblock {\em arXiv:1709.01062}, 2017.

\bibitem{young2018recent}
Tom Young, Devamanyu Hazarika, Soujanya Poria, and Erik Cambria.
\newblock Recent trends in deep learning based natural language processing.
\newblock {\em ieee Computational intelligenCe magazine}, 13(3):55--75, 2018.

\bibitem{zagoruyko2016wide}
Sergey Zagoruyko and Nikos Komodakis.
\newblock Wide residual networks.
\newblock {\em arXiv preprint arXiv:1605.07146}, 2016.

\end{thebibliography}

\end{document}